\documentclass[lettersize,journal]{IEEEtran}
\usepackage{amsmath,amsfonts,amssymb}
\usepackage{algorithmic}
\usepackage{algorithm}
\usepackage{array}
\usepackage[caption=false,font=normalsize,labelfont=sf,textfont=sf]{subfig}
\usepackage{textcomp}
\usepackage{stfloats}
\usepackage{url}
\usepackage{verbatim}
\usepackage{graphicx}
\usepackage{cite}
\usepackage{hyperref}
\usepackage{multirow}
\usepackage{booktabs}
\usepackage{color} 
\begin{document}

\title{Efficient RWKV-based Representation Learning for 3D Point Clouds}

\author{Yun Liu, Xuefeng Yan, Liangliang Nan, Xianzhi Li, Peng Li, Zhe Zhu, Honghua Chen, and Mingqiang Wei,~\textit{Senior Member, IEEE}
\thanks{Yun Liu, Peng Li, Zhe Zhu, Honghua Chen, and Mingqiang Wei are with the School of Computer Science and Technology, Nanjing University of Aeronautics and Astronautics, Nanjing, China, and also with the Shenzhen Institute of Research, Nanjing University of Aeronautics and Astronautics, Shenzhen, China (e-mail: yun.liu.lydia@gmail.com, pengl@nuaa.edu.cn, zhuzhe0619@nuaa.edu.cn, chenhonghuacn@gmail.com, mingqiang.wei@gmail.com). }
\thanks{Xuefeng Yan is with the School of Computer Science and Technology, Nanjing University of Aeronautics and Astronautics, Nanjing, China, and also with the Collaborative Innovation Center of Novel Software Technology and Industrialization, Nanjing, China (e-mail: fzjm\_313@nuaa.edu.cn).}
\thanks{Liangliang Nan is with the Urban Data Science section, Delft University of Technology, Delft, Netherlands (e-mail: liangliang.nan@gmail.com). }
\thanks{Xianzhi Li is with the Huazhong University of Science and Technology, Wuhan 430074, China (email: xzli@hust.edu.cn). }
}

\markboth{Journal of \LaTeX\ Class Files,~Vol.~14, No.~8, August~2021}%
{Shell \MakeLowercase{\textit{et al.}}: A Sample Article Using IEEEtran.cls for IEEE Journals}


\maketitle

\begin{abstract}
The recent receptance weighted key value (RWKV) model combines RNN-style recurrence, offering a linear-complexity alternative to Transformers' quadratic self-attention for modeling global dependencies. However, when directly applied to point clouds, RWKV, originally developed for sequential text, struggles to capture local geometric structures and model spatial dependencies effectively. 
To address this, we propose the \textbf{P-RWKV} block, which bridges the gap between sequence modeling and irregular 3D geometry while preserving the efficiency advantages of RWKV. It consists of a Local Perception Expansion (LPE) component to expand contextual perception along the spatio-temporal sequence and a Spatial Context Enhancement (SCE) component to strengthen spatial awareness. 
To validate the effectiveness of P-RWKV for point cloud understanding, we construct PointER, a single-modality self-supervised representation learning framework whose encoder is composed of stacked P-RWKV blocks. Furthermore, we extend P-RWKV to a cross-modality setting and integrate the proposed core sub-modules into multiple architectures, demonstrating strong plug-and-play flexibility and architectural generality. Extensive experiments show that the P-RWKV block and its key sub-modules achieve competitive performance across various tasks with lower computational cost and inference latency. Code will be released upon acceptance.
\end{abstract}


\section{Introduction}
\IEEEPARstart{S}{elf-supervised} representation learning for 3D point clouds has achieved notable success in real-world applications such as autonomous driving, industrial inspection, and embodied intelligence. Among existing approaches, transformer-based architectures~\cite{pang2022masked,zhang2022pointM2AE,chen2023pointgpt} are the dominant paradigm, benefiting from carefully designed pretext tasks and global dependency modeling via self-attention. 
\begin{figure}[h]
    \centering
    \includegraphics[width=1.0\linewidth]{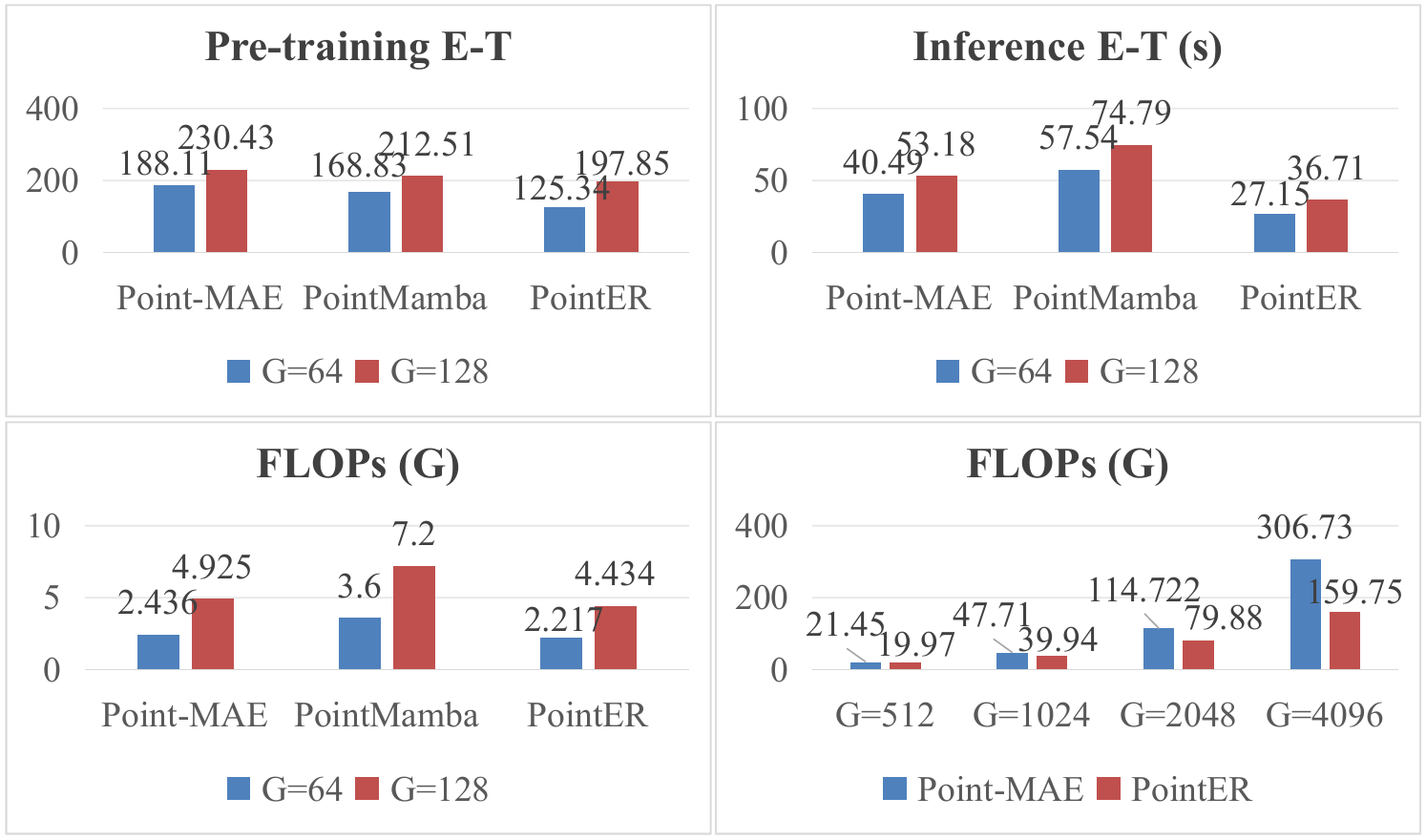} 
    \caption{\textbf{Efficiency comparison under identical settings.}
    We evaluate transformer-based Point-MAE~\cite{pang2022masked}, mamba-based PointMamba~\cite{PointMamba_NeurIPS24}, and the proposed RWKV-based PointER in terms of floating-point operations (`FLOPs'), as well as per-epoch pre-training and inference times (`Pre-train E-T' and `Inference E-T'), with group numbers set to 64 and 128. We further analyze the FLOPs of the inference models of Point-MAE and PointER under input token numbers of 512, 1024, 2048, and 4096. The results indicate that PointER requires lower FLOPs and training/inference costs compared with Point-MAE and PointMamba. Moreover, the FLOPs of PointER exhibit an approximately linear increase with the number of input tokens, consistent with the theoretical computational complexity of RWKV.
}
    \label{fig:Intro_Efficient}
\end{figure}

\par Despite their strong performance, Transformers suffer from quadratic computational and memory costs, which is particularly problematic for point clouds that often contain thousands of discrete and unordered points. This limitation restricts their applicability in large-scale or resource-constrained scenarios and motivates the search for more efficient architectures that preserve representation quality while reducing computational overhead. 
To tackle Transformers' efficiency bottlenecks, new architectures like Mamba~\cite{Mamba_CoRR23} and RWKV~\cite{RWKV_EMNLP23} have recently been proposed. 
Mamba is a linear-complexity sequence modeling architecture built upon structured state space models. 
Its effectiveness has been demonstrated in visual representation learning~\cite{PointMamba_NeurIPS24, Mamba3D_ACM_MM_24, VMamba_NeurIPS24, ViM_ICML24}, achieving performance comparable to or better than Transformers while requiring lower computational resources.
However, its wall-clock efficiency at common resolutions is often comparable to, or even worse than, that of Vision Transformers~\cite{ViG_AAAI25}. 
In contrast, RWKV~\cite{RWKV_EMNLP23, RWKV_Matrix_Valued_CoRR24} provides a hardware-efficient alternative that unifies RNN-style sequential processing with transformer-style parallelizable computation, achieving linear complexity without sacrificing performance. 
Originally developed for natural language processing, RWKV has recently been extended to vision tasks~\cite{RWKV_SAM_24, RWKV_CLIP_24, LSTM_CrossRWKV_24, Vision_RWKV_ICLR25}, achieving transformer-level performance at substantially lower computational cost. 
In 3D point clouds, PointRWKV~\cite{PointRWKV_AAAI25} is an early attempt to adapt RWKV for point cloud representation learning. It directly adopts RWKV for global feature modeling and introduces an additional graph-based module to capture local geometric structures. However, this module stacks multiple MLP layers in a sequential and iterative manner, resulting in increased computational complexity and consequently higher training and inference latency. 

\par In this context, RWKV remains largely unexplored in 3D point cloud processing. This gap stems from several fundamental challenges. 
First, RWKV is inherently designed for serialized text sequences and lacks explicit mechanisms to capture local geometric structures essential for point cloud understanding. 
Second, the causal WKV operation is ill-suited to unordered point tokens, leading to suboptimal global context modeling. 
Third, RWKV does not explicitly encode spatial awareness, which is critical for reasoning over discrete point sets and their geometric relationships. 

\par To overcome these limitations, we propose P-RWKV, which incorporates the RWKV architecture into point cloud representation learning, enabling efficient contextual modeling with linear computational complexity.
Specifically, each P-RWKV block consists of a spatial-mix module and a channel-mix module. The spatial-mix module performs linear-complexity global attention, while the channel-mix module facilitates feature interaction across channels for each token. 
To explicitly capture local geometric structures and enhance spatial perception, we introduce two key components, a Local Perception Expansion (LPE) module and a Spatial Context Enhancement (SCE) module. LPE facilitates feature exchange among neighboring tokens along the spatio-temporal dimension, enriching local geometric information. Moreover, we adopt a bidirectional WKV (Bi-WKV) mechanism from Vision-RWKV~\cite{Vision_RWKV_ICLR25} to model global dependencies in both forward and backward directions in an RNN-like manner while preserving linear complexity. SCE employs a gating mechanism to augment tokens containing spatial neighborhood information and fuses them with the global contextual outputs produced by Bi-WKV.
Collectively, these designs are achieved without introducing additional computational modules, allowing P-RWKV to preserve the computational efficiency of RWKV. Furthermore, the P-RWKV block and its constituent sub-modules can be integrated into existing architectures with minimal architectural modifications.

\par To comprehensively evaluate the effectiveness of the proposed P-RWKV, we construct PointER, a self-supervised representation learning framework built upon stacked residual P-RWKV blocks under a masked autoencoder paradigm. Following standard evaluation protocols, PointER is evaluated on 3D completion, shape classification, and fine-grained segmentation tasks. The experimental results show that PointER achieves competitive performance relative to both mamba-based and transformer-based models, with lower FLOPs and the lowest inference latency, as illustrated in Fig.~\ref{fig:Intro_Efficient}. 
Additionally, we extend P-RWKV and its core components to multiple architectures and learning paradigms under diverse experimental settings to evaluate their applicability across different architectural designs. 
These results highlight the potential of P-RWKV for computation-constrained scenarios such as embodied AI and edge computing. 

\section{Related Work} \label{sec:relatedWork}
\subsection{Models for Point Clouds Analysis}
Unlike conventional representation learning methods based on structured data such as text or images, point clouds are a form of visual data composed of unordered sets of points. Learning robust representations from these discrete and irregular point sets presents unique challenges. Existing approaches are broadly categorized into three classes: \textbf{projection-based methods}, \textbf{voxel-based methods}, and \textbf{point-based methods}. 

\par \textbf{Projection-based methods~\cite{PointPillars_CVPR19}} project irregular point clouds onto various 2D views (e.g., images or depth maps) and then utilize mature 2D backbones for feature extraction. 
Similarly, \textbf{voxel-based methods~\cite{maturana2015voxnet}} convert point clouds into 3D voxel grids, enabling the use of standard convolutional operations. While effective, these methods introduce information loss and discretization artifacts due to their reliance on an intermediate representation. 
\par In contrast, \textbf{point-based methods~\cite{qi2017pointnet, qi2017pointnet++, wang2019DGCNN, zhao2021point, PointTransV2_NeurIPS_22, PointTransV3_CVPR_24, yu2022pointBERT, pang2022masked, zhang2022pointM2AE, chen2023pointgpt}} operate directly on the raw point data, avoiding such preprocessing steps. 
Pioneered by PointNet~\cite{qi2017pointnet}, this family of approaches processes each point individually using multilayer perceptrons (MLPs) before applying a symmetric function to aggregate global features. PointNet++~\cite{qi2017pointnet++} built upon this framework by incorporating hierarchical sampling and grouping strategies to capture local geometric structures across multiple scales.
Following the remarkable success of Transformers~\cite{vaswani2017attention} in natural language processing~\cite{devlin2018bert, radford2019language} and computer vision~\cite{image1616_ICLR_21}, transformer-based point cloud methods~\cite{zhao2021point, PointTransV2_NeurIPS_22, yu2022pointBERT, pang2022masked, zhang2022pointM2AE} has emerged. By leveraging self-attention, these methods effectively capture long-range dependencies and global contextual relationships among points, yielding strong performance across a variety of point cloud understanding tasks.
\par However, both the memory footprint and computational complexity of these models scale quadratically with the input sequence length. The intrinsic complexity of the self-attention mechanism imposes substantial constraints on the applicability of transformer blocks, particularly in time-sensitive and resource-constrained scenarios. 
To address this limitation, recent studies have explored more computationally efficient architectures, such as Mamba~\cite{Mamba_CoRR23} and RWKV~\cite{RWKV_EMNLP23}.

\subsection{Mamba-based Models}
Mamba~\cite{Mamba_CoRR23}, a deep learning model based on State Space Models (SSMs), offers improved computational efficiency and scalability through selective retention or discarding of input information. Following its success in natural language processing~\cite{BlackMamba_NLP_24, MoE_Mamba_NLP24}, recent studies have extended Mamba to 2D images~\cite{MobileMamba_CVPR25, VMamba_NeurIPS24, MambaAD_NeurIPS24} and videos~\cite{Vivim_Video24}, achieving competitive performance. These approaches are widely regarded as a significant step toward achieving linear computational complexity within the vision domain.

More recently, several studies, including PCM~\cite{PCM_AAAI25}, PointMamba~\cite{PointMamba_NeurIPS24}, and Mamba3D~\cite{Mamba3D_ACM_MM_24}, have explored adapting Mamba to point clouds. By representing point cloud data as state vectors, these models can effectively capture the dynamic characteristics of the data, thereby yielding competitive performance. 
However, despite their theoretical linear complexity, the practical wall-time efficiency of these methods is often comparable to, or even inferior to, that of transformer-based methods under identical settings.
\subsection{RWKV-based Models}
Another linear complexity model, Receptance Weighted Key Value (RWKV)~\cite{RWKV_EMNLP23, RWKV_Matrix_Valued_CoRR24}, integrates the efficient and parallelizable training paradigm of Transformers with the low-latency, step-wise inference capabilities of recurrent neural networks (RNNs). 
RWKV initially demonstrated impressive performance in natural language processing~(NLP). Numerous studies have recently explored the adaptation of RWKV to the vision domain~\cite{LineRWKV_24, RWKV_SAM_24, Vision_RWKV_ICLR25, LSTM_CrossRWKV_24, RWKV_CLIP_24}. Among these, Vision-RWKV constructs its encoder by stacking multiple VRWKV blocks in a manner analogous to a Vision Transformer (ViT)~\cite{image1616_ICLR_21}. Extensive experiments demonstrate that Vision-RWKV achieves superior performance compared to ViT, while incurring a slightly reduced computational cost. 

\par When dealing with discrete and unordered point clouds, the application of RWKV remains underexplored. A recent study, PointRWKV~\cite{PointRWKV_AAAI25}, addresses this with a multi-component design. It employs Point-M2AE~\cite{zhang2022pointM2AE} as the backbone for multi-scale feature extraction, adapts RWKV for global modeling, and introduces an additional Local Graph-based Merging (LGM) module to capture local geometric structures.
\par In contrast to prior methods, we propose P-RWKV, which is designed to extract effective features from discrete and unordered point clouds. The P-RWKV block consists of two main components: a spatial-mix module for linear-complexity global attention and a channel-mix module for feature fusion along the channel dimension. 
To address the challenges of learning representations from unstructured point data, the spatial-mix module integrates two key sub-modules: the Local Perception Expansion (LPE) module, which facilitates the exchange of information among neighboring tokens across the spatio-temporal dimension to broaden local perception, and the Spatial Context Enhancement (SCE) module, which reinforces spatial awareness by incorporating local features.


\par Unlike previous works validated on a single model, we conduct a fine-grained evaluation to validate architectural generality. 
We evaluate the effectiveness of P-RWKV for representation learning using a unimodal model, PointER, and a multimodal model, PointER-MM, both built upon stacked P-RWKV blocks. In addition, we systematically transplant the sub-modules of P-RWKV into diverse architectures, demonstrating their plug-and-play flexibility and broad applicability.

\section{Methodology} \label{sec:Method}
Sec.~\ref{sec:Preliminary} briefly reviews the original RWKV model. Sec.~\ref{sec:PRWKV} introduces the proposed P-RWKV architecture, while Sec.~\ref{sec:PointER} presents PointER, a pre-training framework built upon stacked residual P-RWKV blocks for representation learning.
\subsection{Preliminaries}\label{sec:Preliminary}
\par
\begin{figure}[htbp]
    \centering
    \includegraphics[width=0.9\linewidth]{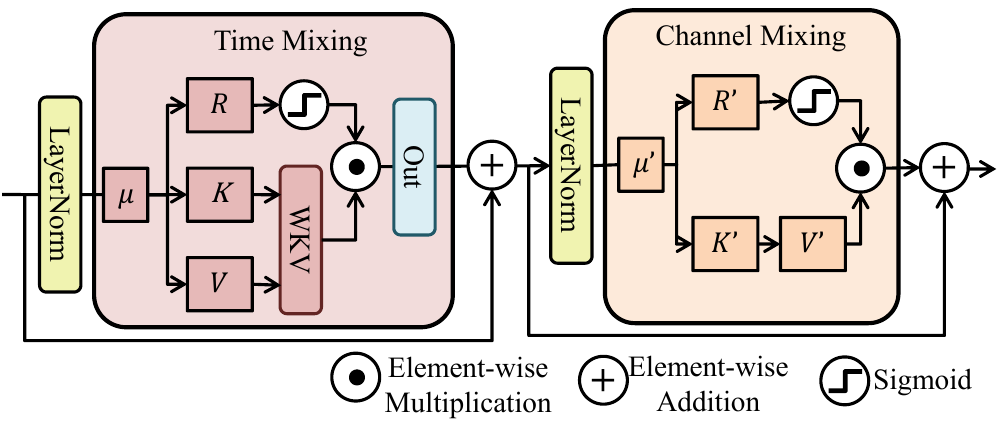}
    \caption{\textbf{Visualization the details of the RWKV block from ~\cite{RWKV_EMNLP23}}. 
}
    \label{fig:RWKV_in_NLP}
\end{figure}

 The RWKV model consists of stacked residual blocks (Fig.~\ref{fig:RWKV_in_NLP}), each containing a time-mixing module and a channel-mixing module. Both modules incorporate recurrent structures to effectively leverage information from previous tokens. 
\par In the \textbf{time-mixing} module, the linear projection vectors $R$, $K$, and $V$ are obtained through linear interpolation between the current and previous timestep inputs, enabling token shift–based information exchange. The WKV operator is then employed to compute attention scores and update hidden states, where a channel-wise vector $W$ is adjusted by relative positional information to serve as a weighting factor. Finally, the sigmoid of the receptance is employed as the output gate. The output of the time-mixing operation is normalized via layer normalization before being fed into the channel-mixing module. 
In the \textbf{channel-mixing} module, the projection vectors $R'$ and $K'$ are similarly obtained through linear interpolation between the current and previous timestep inputs, while output gating is implemented using the sigmoid-activated receptance.

\subsection{P-RWKV}\label{sec:PRWKV}
While RWKV~\cite{RWKV_EMNLP23} has demonstrated remarkable performance in serialized text processing, its direct application to 3D point clouds is not straightforward. The discrete and unordered nature of point clouds presents a significant challenge for capturing local and global features. Unlike the sequential text data where RWKV leverages token shift in the temporal dimension to exchange information with tokens from the previous timestep, the point cloud domain requires a dedicated local feature expansion module to enable token interaction across spatio-temporal dimensions. Moreover, when extracting global features, RWKV exhibits insufficient spatial awareness for point cloud data, necessitating an enhanced spatial context enhancement module. 

\par To address these limitations while preserving the linear complexity and efficiency of RWKV, we propose a general P-RWKV block. Each block consists of a spatial-mix module and a channel-mix module to process input tokens derived from point embeddings and positional embeddings. 
The spatial-mix module functions as a linear-complexity global attention mechanism, effectively capturing long-range dependencies across temporal and spatial positions, while the channel-mix module acts as a feed-forward network (FFN) to facilitate feature interaction along the channel dimension. 
\par \textbf{Spatial-Mix:} We design two specialized sub-modules within the spatial-mix component: the Local Perception Expansion (LPE) module and the Spatial Context Enhancement (SCE) module. 
The LPE module captures contextual features along the spatio-temporal sequence, while the SCE module further enhances spatial context cues, thereby effectively adapting RWKV for point cloud understanding.

\begin{figure}[htbp]
    \centering
    \includegraphics[width=1.0\linewidth]{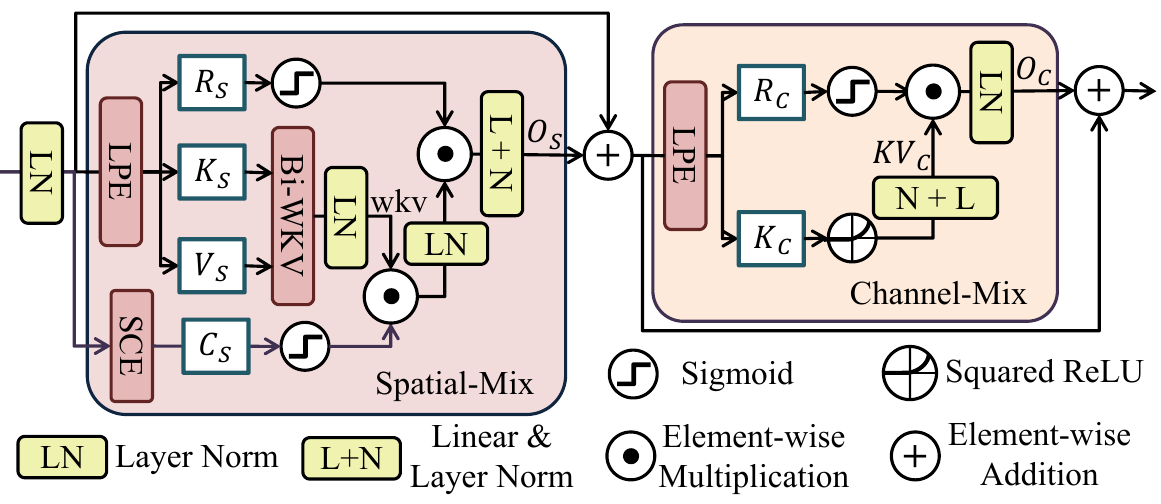}
    \caption{\textbf{Visualization of the details of the P-RWKV block}, which is composed of a spatial-mix module and a channel-mix module. 
    `Bi-WKV' is a bidirectional RNN cell that serves as an efficient alternative to global attention by capturing contextual information in both forward and backward directions. 
    The spatial-mix module is designed to capture local geometric structures by modeling spatial dependencies among neighboring points, while the channel-mix module aggregates and transforms feature representations across channels.
}
    \label{fig:PRWKV_block}
\end{figure}
\par Fig.~\ref{fig:PRWKV_block} presents a detailed illustration of the proposed P-RWKV block for point clouds. Notably, both the inputs and the information flow differ not only between blocks but also within individual blocks. 
\par Specifically, in the first block, the input to the LPE module is denoted as $X^E$, which is obtained by summing the group embedding tokens $T^E_v$ with their corresponding positional encodings $L^E_v$: 
\begin{equation}
\begin{split}
X^E=T^E_v+L^E_v,
\label{eq:Spatial_Input_LPE}
\end{split}
\end{equation}
where $T^E_v$ alone is fed to the SCE module. 
The input tokens of the $l$-th block are denoted as $X^E_l$ (i.e., $X^E_l = X^E$). We forward them to the $(l+1)$-th block via a residual connection, ensuring the preservation of original features and facilitating gradient propagation. 
\par LPE is implemented as a shift-based offset along the temporal and spatial dimensions within each block, leveraging the PyTorch~\cite{PyTorch_NeurIPS19} library. Serving as a flexible extension of the original token shift operation, LPE is applied at the initial stage of each spatial-mix and channel-mix module. 
\par First, the LPE module indexes the local neighborhood around center points and integrates tokens from the previous timestep, thereby enriching local features across both spatial and temporal dimensions.
This operation enables all tokens to be shifted and linearly interpolated with their neighboring tokens:
\begin{equation}
\begin{split}
LPE_{(*)}(X^E)=\mu_{(*)}\odot X^E+(1-\mu_{(*)})\odot X^\gamma, 
\label{eq:Q-shift}
\end{split}
\end{equation}
where $X^\gamma=Concat(X^1[:,0:C/4],X^2[:,C/4:C/2],X^3[:,C/2:3C/4],X^4[:,3C/4:C])$, $\mu_{(*)}$ is a learnable vector, and $X^1$ and $X^2$ denote the neighboring tokens of $X^E$.
In the spatial-mix module of the first block, $X^3$ and $X^4$ are set equal to $X^1$ and $X^2$. In subsequent blocks, $X^3$ and $X^4$ inherit the values of $X^1$ and $X^2$ from the previous block, enabling the propagation of temporal dependencies. 
In the channel-mix module, $X^1$ and $X^2$ represent the neighbor tokens of the current input $X^E$, while $X^3$ and $X^4$ inherit $X^1$ and $X^2$ from the spatial-mix module. This design enables the incorporation of temporal offsets through LPE along the time dimension. 

\par The subscript $(*)\in \{R,K,V\}$ denotes the three interpolation variants of $X^E$ and $X^\gamma$, controlled by the corresponding learnable vectors $\mu_{(*)}$ for the subsequent calculation of $R$, $K$, and $V$, respectively. Here, `:' represents a slicing operation excluding the end index, and $X^\gamma$ represents the concatenation of the sliced token vectors $X^1, X^2, X^3,$ and $X^4$. 
\par Then, three parallel linear layers are employed to produce the relational matrices $R_s, K_s, V_s\in \mathbb{R}^{T\times C}$: 
\begin{equation}
\begin{split}
R_S=LPE_R(X^E)W^S_R, \\ K_S=LPE_K(X^E)W^S_K, \\ V_S=LPE_V(X^E)W^S_V, 
\label{eq:Spatial_RKV}
\end{split}
\end{equation}
where $T$ denotes the total number of tokens, with $T = v$ in the pre-training stage and $T = g$ during inference.

\par In summary, LPE enables the attention mechanism by incorporating information from both preceding and neighboring tokens with minimal additional computation while maintaining linear computational complexity. Moreover, it expands each token's receptive field, thereby improving contextual coverage in deeper layers.
\par Subsequently, $K_S$ and $V_S$ are fed into Bi-WKV, a bidirectional attention mechanism with linear complexity, to compute the global attention result $wkv \in \mathbb{R}^{T \times C}$: 
\begin{equation}
\begin{split}
wkv=Bi-WKV(K_S,V_S).
\label{eq:Spatial_bi_wkv}
\end{split}
\end{equation}
\par The patch embedding tokens $T^E_v$ represent the embeddings of neighboring points, encapsulating rich spatial relative positional information. 
SCE first performs pooling on $T^E_v$ to obtain a global spatial context, which is then propagated back to all tokens to form $C_S \in \mathbb{R}^{T \times C}$. The resulting $C_S$ is subsequently fused with $wkv$ through element-wise multiplication to enhance spatial awareness: 
\begin{equation}
\begin{split}
E_S=\sigma(C_S)\odot{wkv},\\  C_S=SCE(T^E_v)W_C.
\label{eq:Spatial_E_S}
\end{split}
\end{equation}

\par Finally, the spatio-temporal vector $E_S$ is multiplied by $\sigma(R_S)$, which modulates the output probability of $O_S$: 
\begin{equation}
\begin{split}
O_S=(\sigma(R_S)\odot{E_S})W_O.
\label{eq:Spatial_O_S}
\end{split}
\end{equation}
\par The operator $\sigma$ denotes the sigmoid function, and $\odot$ represents element-wise multiplication. Following the linear projection, the output features are gated by $\sigma$ and combined with $E_S$ via $\odot$, after which layer normalization~\cite{LayerNorm_CoRR2016} is applied for feature stabilization.

\par \textbf{Channel-Mix:} The sum of spatial-mix output tokens $O_S$ and the input tokens $X^E$ is fed into the channel-mix module for channel-wise feature fusion:
\begin{equation}
\begin{split}
X^E=X^E+O_S.
\label{eq:Channel_input}
\end{split}
\end{equation}

The matrices $R_C$ and $K_C$ are obtained in a similar manner as the spatial-mix module:
\begin{equation}
\begin{split}
R_C=LPE_R(X^E)W^C_R, \\ K_C=LPE_K(X^E)W^C_K. 
\label{eq:Channel_RK}
\end{split}
\end{equation}

In the channel-mix module, $V_C$ is obtained by applying a linear projection to $K_C$ after the activation function, modulated by a gating mechanism $\sigma(R_C)$. The final output $O_C$ is then produced through a linear projection of the gated features: 
\begin{equation}
\begin{split}
O_C=(\sigma (R_C)\odot V_C)W_O, \\
where ~V_C=SquaredReLU(K_C)W_V.
\label{eq:Channel_O_C}
\end{split}
\end{equation}

\par Simultaneously, residual connections~\cite{he2016deep} are established from the input tokens to each normalization layer to ensure stable gradient propagation and prevent vanishing gradients in deep networks. Consequently, in subsequent blocks, the input tokens $X^E_{l+1}$ are updated as the sum of the current block's input $X^E_{l}$ and the channel-mix output $O_C$, i.e., $X^E_{l+1} = X^E_{l} + O_C$. 
The output $O_C$ of the final block in the encoder is denoted as $F_v$, which serves as the global feature representation for subsequent processing. 
\subsection{PointER}\label{sec:PointER}
\par
\begin{figure}[h]
    \centering
    \includegraphics[width=1.0\linewidth]{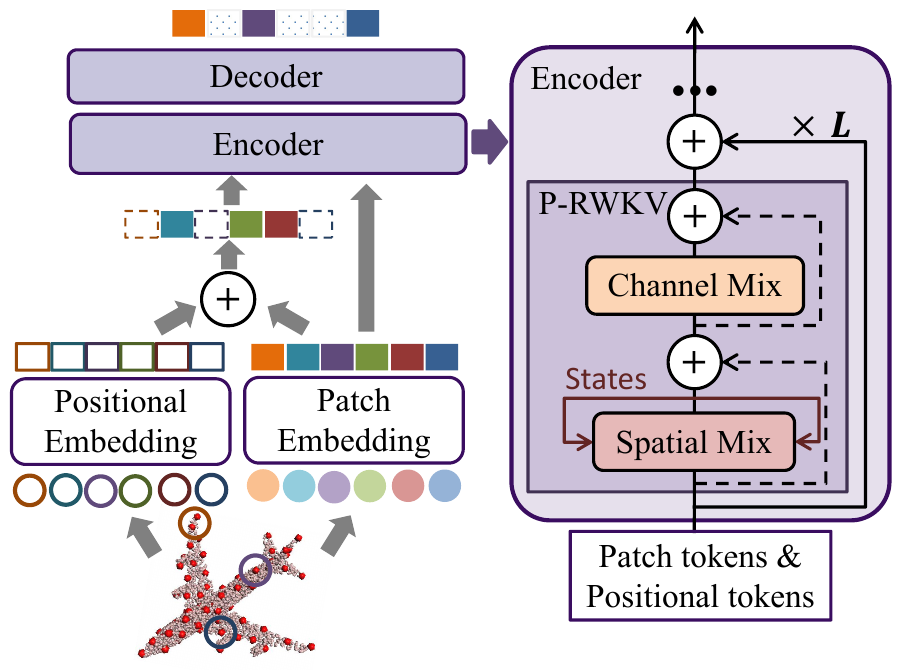}
    \caption{\textbf{Overall architecture of our PointER.}
}
    \label{fig:PointER}
\end{figure}

 Building upon the proposed P-RWKV block, we design a self-supervised learning framework, termed PointER, with Point-MAE as the baseline to validate its representational effectiveness, as illustrated in Fig.~\ref{fig:PointER}.
\par \textbf{Embedding:} Given an input point cloud $P=\{p_{i}|{1}\leq{i}\leq{N}\} \in \mathbb{R}^{N\times3}$, where $p_i=(x_i,y_i,z_i)$, 
we first apply Furthest Point Sampling (FPS) to select $g$ representative centers $C\in\mathbb{R}^{g\times{3}}$. Subsequently, K-Nearest Neighbors (KNN) is employed to retrieve the neighboring points of the centers $C$, forming the patches $Gs\in\mathbb{R}^{g\times{k}\times{3}}$ around the centers. To enhance convergence, all points within each patch are normalized by subtracting their corresponding center coordinates. 

A lightweight PointNet~\cite{qi2017pointnet} is employed to embed the input point patches $Gs$ into d-dimensional tokens $T^E$, serving as the input to the subsequent feature extraction module: 
\begin{equation}
\begin{split}
T^E=PointNet(Gs),  ~T^E\in\mathbb{R}^{g\times{d}} 
\label{eq:eq_Embed}
\end{split}
\end{equation}
where $d$ denotes the embedding dimension of the point tokens.

\par Like Point-MAE, we adopt the commonly used yet effective masked autoencoder architecture to construct the self-supervised pre-training network, PointER. 
Therefore, we randomly mask a subset of centers from $C$ and their corresponding latent tokens from $T^E$, yielding $v$ visible centers $C_v$ and the associated visible latent tokens $T^E_v$. The masked patches $G^{M}$, extracted from $G_s$, are used as prediction targets. 
Subsequently, a learnable multi-layer perceptron embeds the visible centers $C_v$ into positional tokens $L^E_v \in \mathbb{R}^{v \times d}$: $L^E_v = PE(C_v)$.

To enhance pre-training efficiency, both the encoder and the decoder are composed of the proposed P-RWKV blocks.
\par \textbf{Encoder:} 
The encoder consists of $L$ residual P-RWKV blocks, as shown on the right of Fig.~\ref{fig:PointER}. The embedded tokens $T^E_v$ and positional tokens $L^E_v$ are fed into the encoder to obtain latent representations $F_v$ from the visible groups: 
\begin{equation}
\begin{split}
F_v=Encoder(T^E_v,L^E_v),
\label{eq:eq_Encoder}
\end{split}
\end{equation}
where $F_v\in\mathbb{R}^{v\times{d}}$.

\par \textbf{Decoder:} The decoder architecture is similar to the encoder but consists of fewer P-RWKV blocks. The masked centers are embedded into positional tokens, denoted as $L^E_m$. It takes the extracted tokens $F_v$ as inputs and generates point tokens $T_D$ for the following prediction head: 
\begin{equation}
\begin{split}
T_D=Decoder(cat(F_v,T^m),cat(L^E_v,L^E_m)) \\T^m\in\mathbb{R}^{(g-v)\times{d}},  ~T_D\in\mathbb{R}^{(g-v)\times{d}}
\label{eq:eq_Decoder}
\end{split}
\end{equation}
where $L^E_v$ and $L^E_m$ are the positional tokens of visible and hidden centers, respectively. $T^m \in\mathbb{R}^{(g-v)\times{d}}$ represents the initialization of masked tokens by duplicating a learnable masked token of $d$ dimension. 

\par \textbf{Prediction head: } A fully connected (FC) layer is employed as the prediction head to reconstruct the masked patches, which is then reshaped to the original patch dimensions: 
\begin{equation}
\label{eq:Rebuild_maskedPoints}
\begin{split}
\hat{P}=Reshape(FC(T_D)),  ~\hat{P}\in\mathbb{R}^{(g-v)\times{k}\times{3}}
\end{split}
\end{equation}
where $T_D$ are the decoded tokens, and the predicted masked point patches are denoted as $\hat{P}$.

\par \textbf{Loss function: } 
The generation objective for each point cloud is to predict the coordinates of points in the masked patches. Given the predicted point patches $\hat{P}$ and the ground-truth point patches $P^{M}$, the generation loss $\ell$ is defined using both the $\ell_1$-form and $\ell_2$-form of the Chamfer distance~(CD)~\cite{fan2017point}, denoted as $\ell_1$ and $\ell_2$, respectively. 
We employ $\ell=\ell_1+\ell_2$ as the reconstruction loss: 
\begin{equation}
\begin{split}
\ell_{\gamma}=\frac{1}{|\hat{P}|}\sum_{\hat{x}\in{\hat{P}}}\min_{x\in{P^{M}}}||\hat{x}-x||_{\gamma}^{\gamma} \\ +\frac{1}{|P^{M}|}\sum_{x\in{P^{M}}}\min_{\hat{x}\in{\hat{P}}}||\hat{x}-x||_{\gamma}^{\gamma}
\label{eq:loss_Rebuild}
\end{split}
\end{equation}
where ${\gamma}\in \{1,2\}$, $P^{M}\in\mathbb{R}^{(g-v)\times{k}\times{3}}$ denotes the reconstruction target, $|P|$ is the cardinality of the set $P$, and $||\hat{x}-x||_n$ denotes the $\ell_n$ distance between $\hat{x}$ and $x$.

\section{Experiment} \label{sec:Experiments}
We evaluate P-RWKV-based models against existing methods on point cloud completion, shape classification, and fine-grained segmentation, together with theoretical and empirical analyses of computational complexity.
In addition, we extend P-RWKV and its core sub-modules to multiple self-supervised learning paradigms and network architectures, including TAP~\cite{Wang_2023_TAP}, Gated Linear Attention (GLA), and Point Transformer V3 (PTv3)~\cite{PointTransV3_CVPR_24}. The resulting models demonstrate the architectural generality of the proposed modules.
Finally, extensive ablation studies are conducted to analyze the effectiveness of the proposed design choices.

\par For fair and rigorous comparison, we include methods with publicly available implementations and reproduce several of them using their released code under identical settings. 
The reproduced results are denoted as `Rep.'. Besides fine-tuning the pre-trained encoder, we further perform fully supervised training using the encoder of PointER, denoted as `Sup.'.
Performance is evaluated using standard accuracy and voting accuracy~\cite{liu2019relation}. To demonstrate the efficiency advantage of RWKV's linear complexity, we report the number of parameters (`Param.'), floating-point operations (`FLOPs'), and per-epoch time (`E-T') under identical settings.

\par \textbf{Implementation Details}. 
The encoder of PointER consists of $12$ P-RWKV blocks with a hidden dimension of $384$. The model is pre-trained on ShapeNet55~\cite{chang2015shapenet} for 300 epochs with a batch size of 128. Each input point cloud contains 1,024 points and is partitioned into 64 patches unless otherwise specified. AdamW with cosine learning rate decay is adopted, using an initial learning rate of 0.001 and a weight decay of 0.05. 
During fine-tuning, the point cloud is divided into 64 patches. Unless otherwise specified, we follow Point-MAE~\cite{pang2022masked} and adopt the same hyperparameter settings and task heads across all downstream tasks. All experiments are conducted on an NVIDIA RTX 3090 GPU.

\subsection{3D Completion}  \label{sec:Completion}
To evaluate both the reconstruction capability and the cross-dataset transferability of the pre-trained model, we conduct occlusion completion experiments on ShapeNet55~\cite{chang2015shapenet} and ModelNet40~\cite{wu2015_modelNet}. 

\par Unlike the conventional random masking strategy employed during pre-training, we adopt a more challenging consecutive masking scheme for the downstream completion task. Specifically, we mask consecutive groups of points and retain only a portion of each instance as input, with a mask ratio of 60\%. This strategy causes the remaining visible points to lack global structural cues, thereby significantly increasing the difficulty of reconstruction and providing a more rigorous test of the model's robustness. 

\par
\begin{figure}[htbp]
    \centering
    \includegraphics[width=1.0\linewidth]{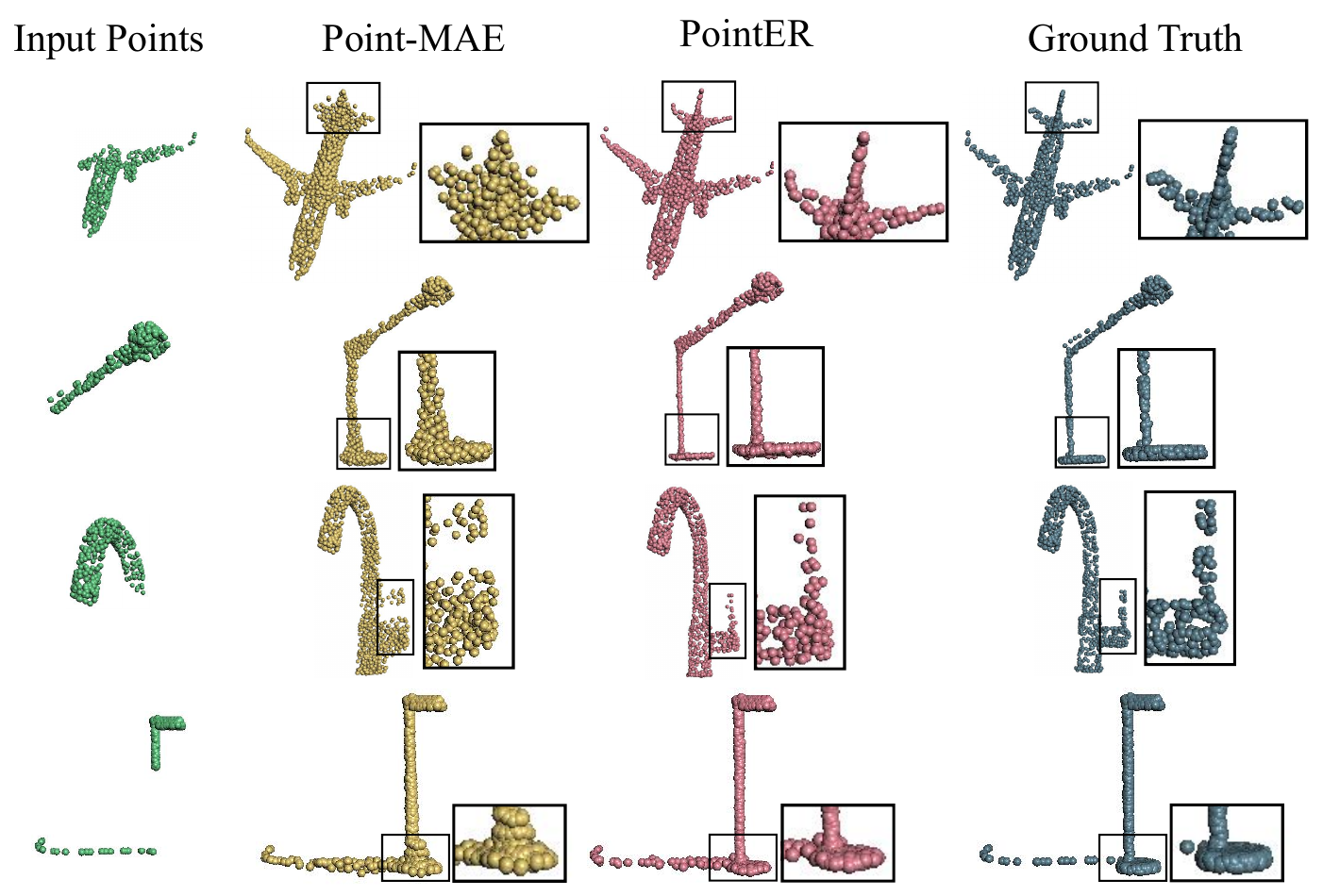}
    \caption{\textbf{Visualization of the completion results on ShapeNet55 with frozen pre-trained models.}
    }
    \label{fig:MaskedRecon_SN55}
\end{figure}

 \textbf{Visualization of completion on ShapeNet55.} 
As shown in the `Input Points' column of Fig.~\ref{fig:MaskedRecon_SN55}, consecutive masking causes severe structural information loss. While both our method and Point-MAE are able to recover the overall shape, our approach produces finer local geometric details that better match the ground truth, resulting in more complete reconstructions.
\par
\begin{figure}[htbp]
    \centering
    \includegraphics[width=1.0\linewidth]{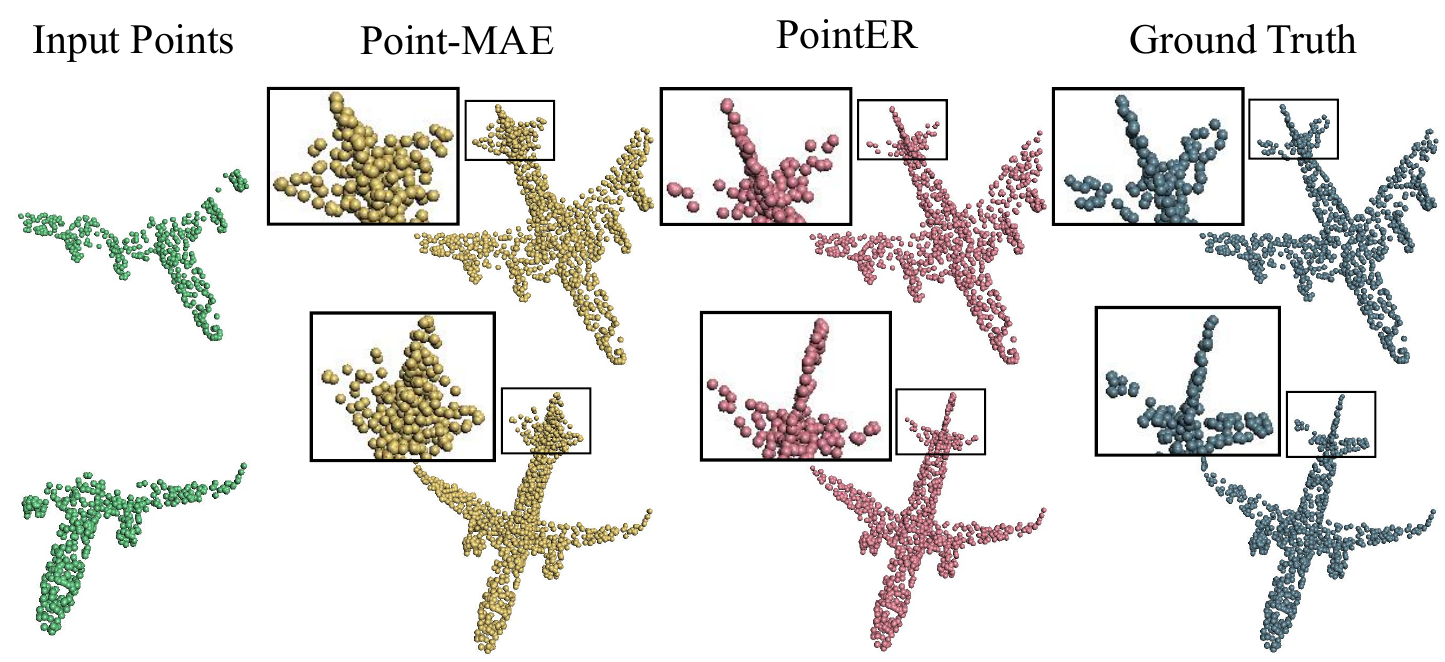}
    \caption{\textbf{Visualization of the completion results on ModelNet40 with frozen pre-trained models.}
    }
    \label{fig:MaskedRecon_MN40}
\end{figure}

 \textbf{Visualization of completion on ModelNet40.} To further evaluate the cross-dataset completion capability of our pre-trained model, we perform a completion task on ModelNet40~\cite{wu2015_modelNet} instances using the model pre-trained on ShapeNet55. As shown in Fig.~\ref{fig:MaskedRecon_MN40}, our method not only reconstructs the overall structure but also infers fine-grained local details and yields smoother edges. In contrast, Point-MAE exhibits limited capability in restoring subtle local features, leading to less accurate reconstructions.

\par \textbf{Quantitative results of completion on ShapeNet55 and ModelNet40.} 
We randomly sample partial point sets from ShapeNet55 and ModelNet40 and perform completion with the frozen pre-trained models. Chamfer Distance (CD) and Earth Mover's Distance (EMD) are employed as quantitative metrics to objectively assess reconstruction fidelity.
\begin{table}[htbp] 
\caption{\textbf{Completion results on ShapeNet55 and ModelNet40. } Note that, all results are multiplied by 1k.}
\centering
\begin{tabular}{lcccc}
\toprule
\multirow{2}{*}{\textbf{Methods}}  & \multicolumn{2}{c}{\textbf{ShapeNet55}} & \multicolumn{2}{c}{\textbf{ModelNet40}} \\ 
\cmidrule(lr){2-3} \cmidrule(lr){4-5}
& CD~$\downarrow$ & EMD~$\downarrow$ & CD~$\downarrow$ & EMD~$\downarrow$ \\ 
\midrule
Point-MAE~(Rep.)   & 2.245   & 6.273   & 2.032   & 4.214\\ 
PointMamba~(Rep.) & 0.975   & 3.954  & 2.031   & 4.468 \\
Mamba3D~(Rep.)   & 1.742   & 5.543  & 2.025   & 8.377 \\
PointER       & 1.306   & 3.326  & 1.971   & 4.117 \\ 
\bottomrule
\end{tabular}
\label{tab:Comp_CD_CMD_SN55_MN40}
\end{table}
 
\par As shown in Tab.~\ref{tab:Comp_CD_CMD_SN55_MN40}, PointER achieves the lowest EMD on ShapeNet55, while exhibiting a slightly higher CD than PointMamba. When pre-trained on ShapeNet55 and transferred to ModelNet40 for completion, PointER achieves the best overall performance among all evaluated methods.
\par In summary, the superior reconstruction of both global structure and local details in this cross-dataset setting demonstrates the effectiveness of the P-RWKV block, equipped with LPE and SCE, as a robust feature extractor.

\subsection{Shape Classification} \label{sec:ShapeClassification}

\par \textbf{Shape classification on ModelNet40~\cite{wu2015_modelNet}.} ModelNet40 consists of $12,311$ 3D CAD models spanning $40$ object categories, partitioned into $9,843$ instances for training and $2,468$ instances for testing. 

\begin{table}[htbp]
\caption{ \textbf{Shape classification on ModelNet40~\cite{wu2015_modelNet}.}
`Trans' denotes Transformer.
}
\centering
\scalebox{0.9}{
\begin{tabular}{lccc}
\toprule
\centering  
Methods                  &Backbone     & Booktitle & Acc.~(Vote)\\ \midrule
PointNet~\cite{qi2017pointnet}    & -  & $CVPR^{17}$    & 89.2   \\
PointNet++~\cite{qi2017pointnet++} & - & $NeurIPS^{17}$ & 90.7  \\
DGCNN~\cite{wang2019DGCNN}         & -   & $TOG^{19}$    & 92.9 \\
PointNeXt~\cite{qian2022pointnext} & -  &$NeurIPS^{22}$ & 94.0   \\
PointMLP~\cite{PointMLP_ICLR22}    & -  &$ICLR^{22}$    & 94.5  \\ 
PointER~(Sup.,~G=64)  & RWKV  & -   &93.34 \\ 
\midrule
Point-BERT~\cite{yu2022pointBERT}  & Trans  & $CVPR^{22}$   & 93.2  \\ 
MaskPoint~\cite{Liu_ECCV_MaskPoint} & Trans & $ECCV^{22}$    & 93.8  \\
Point-MAE~\cite{pang2022masked}   & Trans   & $ECCV^{22}$   & 93.8 \\ 
Point-M2AE~\cite{zhang2022pointM2AE}& Trans  & $NeurIPS^{2022}$   & 94.0 \\ 
PointMamba~(Rep.,~G=64)             & Mamba  & $NeurIPS^{24}$    & 93.12 \\ 
PointMamba~(Rep.,~G=128)            & Mamba  & $NeurIPS^{24}$    & 93.40 \\ 
PCM~\cite{PCM_AAAI25}              & Mamba     & $AAAI^{25}$ & 93.4  \\
Mamba3D~\cite{Mamba3D_ACM_MM_24}   & Mamba  & $ACM~MM^{25}$   & 94.4  \\ 
PointRWKV~(Rep.,~G=64)   & RWKV  & $AAAI^{25}$   & 93.38 \\  
PointRWKV~(P-RWKV,~G=64)   & RWKV  & -   & 93.44\\ 
PointRWKV~(Rep.,~G=128)   & RWKV  & $AAAI^{25}$   & 93.46  \\
PointRWKV~(P-RWKV,~G=128)   & RWKV  & -   & 93.59  \\
Point-MAE~(Rep.,~G=64)                & Trans     & $ECCV^{22}$  & 93.24  \\
PointER~(Point-MAE,~G=64)           & RWKV           & -       & \textbf{93.41} \\ 
Point-MAE~(Rep.,~G=128)            & Trans  & $ECCV^{22}$    & 93.39 \\
PointER~(Point-MAE,~G=128)      & RWKV    & -     & \textbf{93.47}  \\ 
\bottomrule
\end{tabular}}
\label{tab:shapeClsMN40_GL_MLP3}
\end{table}
\par As reported in Tab.~\ref{tab:shapeClsMN40_GL_MLP3}, our method outperforms the baseline Point-MAE on the classification task, achieving an improvement of over $+0.17\%$. Moreover, it delivers strong performance under the fully supervised setting, demonstrating that the encoder, constructed by stacking residual P-RWKV blocks, is highly effective for point cloud classification.
\par 
For fair comparison, we replace the corresponding blocks in PointRWKV with our proposed P-RWKV blocks, yielding a modified variant denoted as PointRWKV (P-RWKV).
Under the same settings, the variant equipped with P-RWKV achieves superior performance on ModelNet40.
\par \textbf{Shape classification on ScanObjectNN~\cite{uy2019revisiting}.} 
This dataset comprises approximately $15,000$ objects across $15$ distinct categories, captured from real-world indoor scenes with cluttered backgrounds. 
As shown in Tab.~\ref{tab:Cls_ScanObjectNN}, PointER achieves higher classification accuracy than Point-MAE~\cite{pang2022masked} on the `OBJ-BG' and `OBJ-ONLY' variants. Under the fully supervised setting, PointER also performs well across all three variants. These results demonstrate its effectiveness and robustness on real-world datasets.
\begin{table}[htbp]
\caption{\textbf{Shape classification on ScanObjectNN~\cite{uy2019revisiting}.} 
}
\centering
\begin{tabular}{cccc} 
\toprule
Methods                            & OBJ-BG & OBJ-ONLY & PB-T50-RS \\ \midrule
PointNet~\cite{qi2017pointnet}    & 73.3   & 79.2     & 68.0  \\ 
PointNet++~\cite{qi2017pointnet++} & 82.3   & 84.3   & 77.9  \\ 
DGCNN~\cite{wang2019DGCNN}         & 82.8   & 86.2   & 78.1 \\ 
Transformer~\cite{yu2022pointBERT}   & 79.86  & 80.55   & 77.24 \\ 
PointER~(Sup.)   & 85.54 & 86.41 & 82.55 \\ 
\midrule
Transformer+OcCo~\cite{wang2021OCCO} & 84.85  & 85.54   & 78.79 \\ 
Point-BERT~\cite{yu2022pointBERT}   & 87.43   & 88.12   & 83.07  \\ 
MaskPoint~\cite{Liu_ECCV_MaskPoint}   & 89.3    & 89.7  & 84.6  \\ 
PointMamba~\cite{PointMamba_NeurIPS24}  & 90.71   & 88.47   & 84.87  \\ 
Point-MAE~\cite{pang2022masked}     & 90.02   & 88.29   & \textbf{85.18}  \\ 
PointER~(Point-MAE)         & \textbf{90.38}  & \textbf{88.49}  & 84.93 \\ 
\bottomrule
\end{tabular}
\label{tab:Cls_ScanObjectNN}
\end{table}

\begin{table*}[h]
\caption{\textbf{Part segmentation results on ShapeNetPart~\cite{yi2016scalable}.} 
}
\centering
\scalebox{0.81}{
\begin{tabular}{c|cc|cccccccccccccccc}
\toprule
Methods         & $mIoU_C$   & $mIoU_I$   & aero   & bag    & cap    & car           & chair    & ear      & guitar     & knife      & lamp      & laptop & motor & mug   & pistol & rocket & skate   & table \\ \midrule
PointNet~\cite{qi2017pointnet}     & 80.39   & 83.7    & 83.4    & 78.7       & 82.5    & 74.9          & 89.6  & 73.0       & 91.5        & 85.9          & 80.8      & 95.3   & 65.2  & 93    & 81.2   & 57.9   & 72.8    & 80.6  \\
PointNet++~\cite{qi2017pointnet++}     & 81.85   & 85.1    & 82.4    & 79.0       & 87.7     & 77.3          & 90.8  & 71.8       & 91.0        & 85.9           & 83.7      & 95.3   & 71.6  & 94.1  & 81.3   & 58.7   & 76.4   & 82.6  \\
DGCNN~\cite{wang2019DGCNN}     & 82.33   & 85.2    & 84.0     & 83.4      & 86.7     & 77.8          & 90.6  & 74.7       & 91.2        & 87.5           & 82.8      & 95.7   & 66.3  & 94.9  & 81.1   & 63.5   & 74.5    & 82.6  \\ 
\textcolor{black}{PointER~(Sup.)}   & 82.43 & 85.16 & 83.8 & 82.2 & 81.0 & 79.4 & 90.8 & 76.2 & 91.0  & 87.6 & 84.5 & 95.7 & 72.4 & 94.0 & 84.4 & 59.8 & 74.9 & 81.4\\ 
\midrule
PointMamba~(Rep.)     & 81.98   & 85.27   & 83.7     & 79.8      & 83.8     & 78.6           & 90.7  & 75.2      & 90.9       & 86.5            & 83.9      & 95.7   & 68.1  & 94.2  & 83.6   & 59.2   & 75.1    & 82.6  \\  
Mamba3D~(Rep.)     & 82.56   & 85.35    & 83.6     & 82.7      & 80.9     & 79.6           & 90.9  & 76.2      & 91.3       & 88.0            & 84.1      & 95.7   & 71.6  & 94.0  & 84.2   & 60.9   & 75.1    & 82.0  \\ 
Transformer~\cite{yu2022pointBERT}     & 83.42   & 85.1    & 82.9     & 85.4      & 87.7     & 78.8           & 90.5  & 80.8      & 91.1       & 87.7            & 85.3      & 95.6   & 73.9  & 94.9  & 83.5   & 61.2   & 74.9    & 80.6  \\
Point-BERT~\cite{yu2022pointBERT}       & 84.11  & 85.6    & 84.3     & 84.8        & 88.0    & 79.8          & 91.0  & 81.7      & 91.6       & \textbf{87.9}  & 85.2 & 95.6   & 75.6      & 94.7    & 84.3     & 63.4     & 76.3       & 81.5  \\
\textcolor{black}{PointRWKV~(Rep.)}      & 84.01  & 85.70    & 84.4     & 84.6        & 86.8    & 79.8       & 91.0  & 80.5      & 92.0       & 87.4  & 85.7 & 95.6   & 76.0      & 94.8    & 84.7     & 63.4     & 76.1       & 81.1  \\
\textcolor{black}{PointRWKV~(P-RWKV)}     & 84.26  & 86.50    & 84.7     & 86.1        & 86.9    & 80.7        & 91.2  & 80.5      & 92.0       & 87.6  & 85.6 & 95.8   & 76.3      & 94.4    & 84.6     & 63.5     & 76.1       & 82.2  \\
Point-MAE~\cite{pang2022masked}         & 84.19    & 86.1     & 84.3      & \textbf{85.0}      & 88.3           & 80.5        & 91.3   & \textbf{78.5}      & \textbf{92.1}    & 87.4  & \textbf{86.1}    & \textbf{96.1}   & 75.2  & 94.6   & 84.7      & \textbf{63.5}  & 77.1    & 82.4  \\
PointER   & \textbf{84.23}  & \textbf{86.44}  & \textbf{84.7} & 84.2  & \textbf{88.8}  & \textbf{80.9}  & \textbf{91.5}   & 73.7   & 92.0   & 87.1     & 85.6     & 96.0   & \textbf{77.6}  & \textbf{95.6}  & \textbf{85.8}    & \textbf{64.0}     & \textbf{77.4}      & \textbf{82.8}  \\  
\bottomrule
\end{tabular}}
\label{tab:partSegDetail}
\end{table*}
\begin{figure*} [htbp] 
    \centering
    \includegraphics[width=1.0\linewidth]{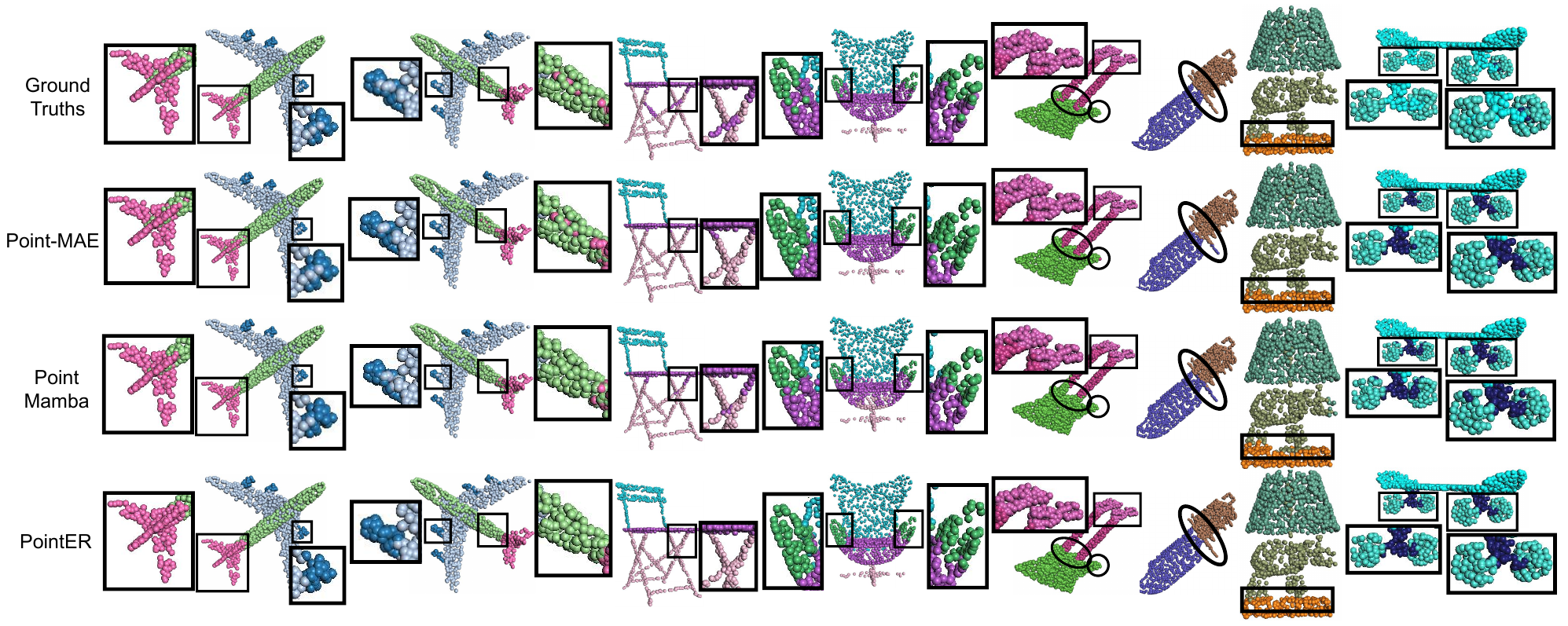}
    \caption{\textcolor{black}{\textbf{Visualization of the part segmentation results on the test set of ShapeNetPart~\cite{yi2016scalable}.}} 
    }
	  \label{fig:multiClass_partSeg}
\end{figure*}
\subsection{Part Segmentation} \label{sec:PartSeg}
\par The task of part segmentation aims to assign fine-grained class labels to each point within an instance. We conduct our part segmentation experiments on the ShapeNetPart dataset~\cite{yi2016scalable}, which consists of $16,881$ samples from $16$ categories, annotated with a total of $50$ distinct parts. 
The input point clouds are uniformly sampled into $2,048$ points. We adopt the segmentation head from Point-MAE~\cite{pang2022masked}, which concatenates the extracted features from the encoder layers. 
\par \textcolor{black}{For fair comparison, we reproduce the publicly available implementations of PointMamba~\cite{PointMamba_NeurIPS24}, Mamba3D~\cite{Mamba3D_ACM_MM_24}, and PointRWKV~\cite{PointRWKV_AAAI25}. In addition, we replace the corresponding blocks in the PointRWKV encoder with our proposed P-RWKV blocks under identical settings.} As shown in Tab.~\ref{tab:partSegDetail}, our method achieves competitive performance in fine-grained part segmentation across most categories. 
\textcolor{black}{
Furthermore, compared with PointRWKV, the model equipped with our proposed P-RWKV blocks achieves superior performance. The consistent improvements in fine-grained segmentation further demonstrate the effectiveness of P-RWKV in capturing local geometric details.}

\par \textcolor{black}{\textbf{Visualization of part segmentation on ShapeNetPart.}} 
Fine-grained part segmentation is a fundamental task in 3D vision with substantial practical value, as it facilitates a more detailed understanding of object geometry and functional components. 
We visualize part segmentation results on representative categories and compare them with transformer- and mamba-based methods (Point-MAE and PointMamba). 
\par \textcolor{black}{As shown in Fig.~\ref{fig:multiClass_partSeg}, when segmentation errors exist in the ground truth (e.g., the last column), models based on the pre-trained encoder are able to produce reasonable predictions by leveraging the knowledge accumulated during pre-training. Moreover, our approach more precisely delineates the boundaries between adjacent parts, particularly in complex edge regions. This highlights the efficacy of our designed local perception module in capturing fine-grained geometric details.}
\begin{figure} [htbp] 
    \centering
    \includegraphics[width=1.0\linewidth]{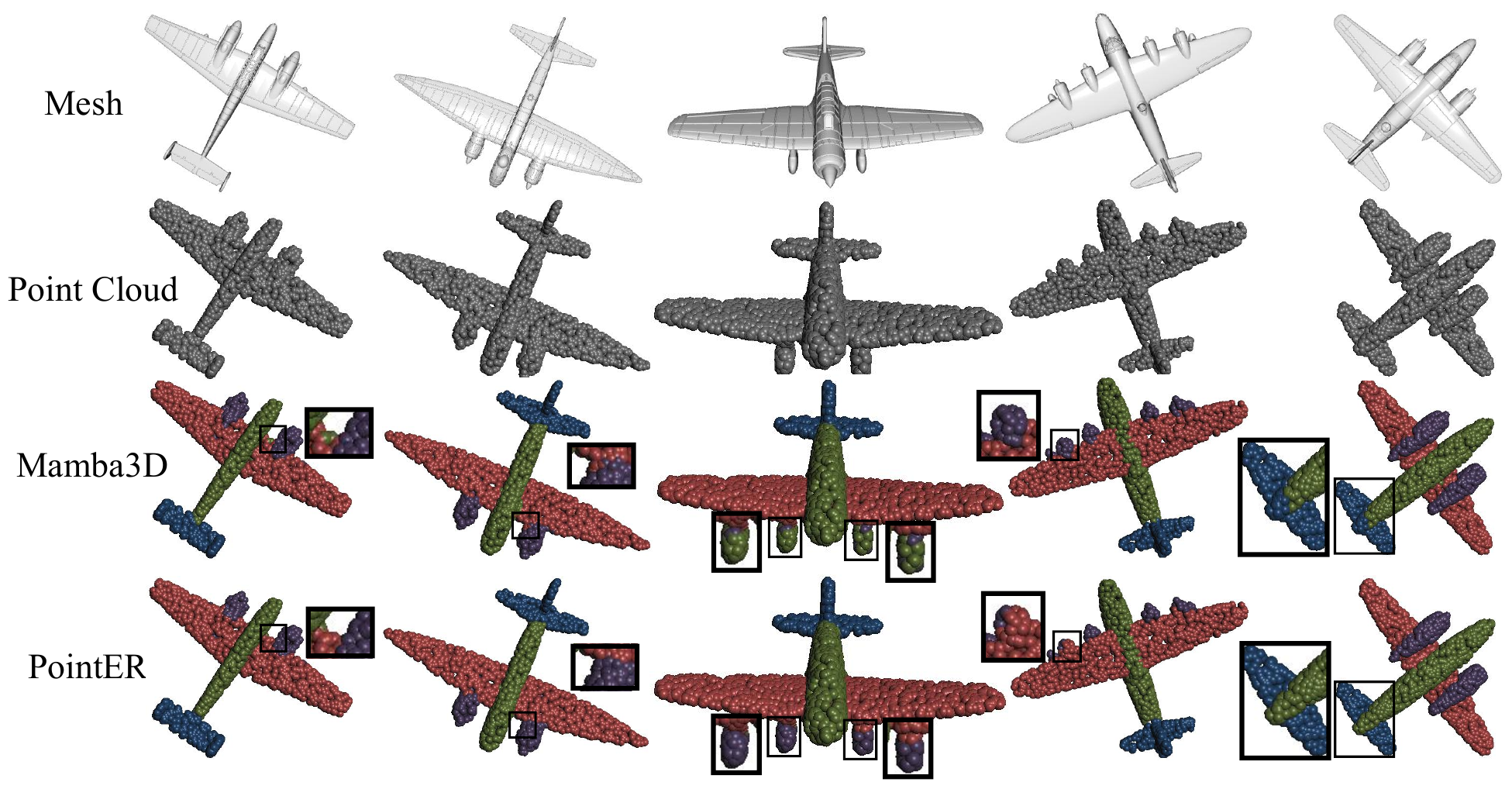}
    \caption{\textcolor{black}{\textbf{Visualization of part segmentation results on our collected dataset.}}}
    \label{fig:PartSeg_realworld_Air}
\end{figure}
\par \textcolor{black}{\textbf{Visualization of part segmentation on our collected dataset.} We collect a large number of real-world aircraft models in STL format, convert them into point clouds, and uniformly downsample them to 2,048 points. Our method is applied to perform part segmentation. Five representative examples are shown in Fig.~\ref{fig:PartSeg_realworld_Air}, demonstrating strong effectiveness and generalization on real-world data.}
\par \textcolor{black}{The first row shows the original mesh models, and the second row presents the corresponding downsampled point clouds. We compare PointER with Mamba3D on a real-world dataset. While both methods generalize well, our approach yields more accurate segmentation, especially along object boundaries and in fine-grained local structures.}
\begin{table*}[h]
\caption{\textbf{Semantic segmentation results on S3DIS~\cite{S3DIS_2016CVPR}.} 
}
\centering
\scalebox{0.89}{
\begin{tabular}{c|ccc|ccccccccccccc}
\toprule
Methods                         & $oAcc$       & $mAcc$         & $mIoU_C$     & ceiling       & floor           & wall    & beam & column & window & door & table & chair & sofa  & bookcase  & board  & clutter  \\ \midrule
PointMamba   & 89.65     & 83.08      & 73.78   & 95.7    & 96.1    & 77.8    & 82.0    & 73.4    & 82.6    & 79.8    & 70.3   &  78.6   & 51.4    & 56.5   &  49.8   &  65.2 \\ 
Mamba3D   & 90.19     & 83.42      & 74.45   & 95.8    & 98.0    & 78.3    & 82.8    & 72.6    & 82.5    & 81.0    &  71.8   & 82.1   & 51.2    & 55.8   &  50.4   &  65.8 \\ 
Point-MAE  & 90.29       & 82.18         & 73.14          & 94.4          & 96.8            & 77.6     & \textbf{82.0}  & 68.5          & 78.4            & 79.0       & \textbf{73.3}             & \textbf{82.9}        & 45.4             & 52.6      & \textbf{55.5}   & 64.6 \\
PointER       &\textbf{90.31} &\textbf{84.37} & \textbf{75.11}   & \textbf{96.1} & \textbf{97.5}  & \textbf{78.0}   & 78.9    & \textbf{76.7}  & \textbf{80.3}   & \textbf{80.6}  & 72.9   & 82.3  & \textbf{57.2} & \textbf{55.9} & 52.6       & \textbf{67.3}  \\
\bottomrule
\end{tabular}}
\label{tab:semanticSeg_Class}
\end{table*}
\subsection{Semantic Segmentation} \label{sec:SemSeg}
\par \textcolor{black}{Real-world scenes present significant challenges due to their complexity, including noise, occlusions, and structural variations. We evaluate on fine-grained semantic segmentation with the Stanford Large-Scale 3D Indoor Spaces (S3DIS) dataset~\cite{S3DIS_2016CVPR}. S3DIS consists of 3D point clouds collected from $272$ rooms across $6$ indoor areas, where each point is annotated into one of $13$ semantic categories. 
} 
\par We fine-tune the pre-trained model on Areas 1–5 ($20,291$ scenes) and evaluate per-category segmentation results on the held-out Area 6 ($3,294$ scenes). 
As summarized in Tab.~\ref{tab:semanticSeg_Class}, our method achieves superior performance on all three aggregated metrics and across most categories, demonstrating the effectiveness of P-RWKV for fine-grained semantic segmentation.

\subsection{Computational Complexity Analysis}
\textcolor{black}{The computational cost of standard transformer-based blocks grows quadratically with the sequence length due to the self-attention mechanism. In contrast, mamba-based blocks achieve linear computational complexity with respect to the sequence length. Similarly, RWKV-based blocks also exhibit linear complexity while maintaining a smaller constant factor, resulting in more efficient computation in practice. }
\textcolor{black}{PointMamba~\cite{PointMamba_NeurIPS24}, Mamba3D~\cite{Mamba3D_ACM_MM_24}, PointRWKV~\cite{PointRWKV_AAAI25}, and the proposed PointER extend Mamba and RWKV to point cloud representation learning. To ensure a fair comparison of computational complexity and inference efficiency, all methods are re-evaluated under identical experimental settings on the same hardware platform. The results are summarized in Tab.~\ref{tab:Efficient_Compare}.}

\par However, RWKV’s global receptive field is insufficient for modeling local point geometry~\cite{PointRWKV_AAAI25}. To mitigate this issue, PointRWKV adopts Point-M2AE as the backbone and introduces an additional Local Graph-based Merging (LGM) module within the PRWKV block to enrich local geometric representations. 
Accordingly, the computational complexity of the PRWKV block comprises two components: the RWKV module directly transferred to point clouds and the additional LGM module. By stacking multiple MLP layers with complexity $O(kNd^2)$, LGM yields an overall complexity of $O(\lambda kNd^2)$, exceeding that of RWKV, $O((A+4r)Nd^2)$, and resulting in higher training and inference latency. 
\par \textcolor{black}{In contrast, in the proposed P-RWKV block, \textbf{LPE} captures contextual perception along the spatio-temporal sequence via a token-shift mechanism based on a precomputed neighbor index matrix, which is computed once and shared across all experiments. Meanwhile, \textbf{SCE} enhances spatial context cues through pooling and propagation. As a result, the P-RWKV block preserves linear complexity $O((A+4r)Nd^2)$ while achieving the lowest inference latency.} 
\begin{table*}[htbp] 
\caption{ 
\textbf{Computational complexity and inference latency comparison.}
The per-epoch inference time (`Inf. E-T') is measured using a global fine-tuning strategy, and the masking ratio during pre-training is set to 0.6.
}
\centering
\begin{tabular}{lcccccccc}
\toprule 
\centering  
\multirow{2}{*}{\textbf{Methods}} &\multirow{2}{*}{\textbf{Backbone}}  & \multicolumn{2}{c}{\textbf{Pre-training Model}} & \multicolumn{4}{c}{\textbf{Classification Model}} \\ 
\cmidrule(lr){3-4} \cmidrule(lr){5-8}
& & P.(M)~$\downarrow$ & FLOPs(G)~$\downarrow$ & ModelNet40~$\uparrow$ & Inf. E-T~(s)~$\downarrow$ & P.(M)~$\downarrow$ & FLOPs(G)~$\downarrow$ \\ 
\midrule
Point-M2AE~\cite{zhang2022pointM2AE}  & Transformer   &15.3 &3.6  &94.0 &-  &-  &- \\
Point-M2AE~(Rep.,~G=[64,32,16])  & Transformer   &38.822 &2.593 & 93.21 & 43.69 &23.879  & 2.311\\  
Point-M2AE~(Rep.,~G=[512,256,64]) & Transformer  &38.822 &19.974  & 93.59 & 135.49 &23.879 &18.530\\ 
PointRWKV~(Rep.,~G=[64,32,16])   & RWKV   &28.675 & 4.278  & 93.38 & 119.630 &64.302 &6.832\\  
PointRWKV~(Rep.,~G=[128,64,32])   & RWKV  &28.675 & 9.873   & 93.46 & 295.457  &64.302 &13.663 \\  
PointRWKV~(Rep.,~G=[512,256,64])   & RWKV  &28.675 & 39.701   & - & -  &64.302 &50.644 \\ 
PointRWKV~(P-RWKV,~G=[64,32,16])   & RWKV &16.276 & 1.719   & 93.44 & 39.254 & 25.286  & 4.318 \\ 
PointRWKV~(P-RWKV,~G=[128,64,32])   & RWKV&16.276 & 3.437    & 93.59 & 65.426 & 25.286 & 8.635\\
PointRWKV~(P-RWKV,~G=[512,256,64])   & RWKV &16.276 & 13.725   & - & -  & 25.286 &32.040\\ \midrule
Point-MAE~(Rep.,~G=64)            & Transformer  &29.006 &2.042  & 93.24  &40.49 &22.1 &2.436\\
Point-MAE~(Rep.,~G=128)           & Transformer   &29.006 & 4.122  & 93.39 & 53.18 &22.1 & 4.925\\  
PointMamba~(Rep.,~G=64)             & Mamba  &16.069 &-  & 93.12 & 57.54 &12.3 &3.6\\ 
PointMamba~(Rep.,~G=128)            & Mamba  &16.069 &- & 93.40 & 74.79 &12.3 &-\\ 
Mamba3D~(Rep.,~G=64)   & Mamba    &23.842 & 1.864  & - & 63.52 & 16.934 & 1.936\\
Mamba3D~(Rep.,~G=128)   & Mamba    &23.842 & 3.742  & 94.4 & 90.476  & 16.934  & 3.861\\
PointER~(Point-MAE,~G=64)         & RWKV   &33.218 &1.962  & \textbf{93.41} &27.15 &25.701 &2.217\\ 
PointER~(Point-MAE,~G=128)      & RWKV   &33.218 &3.925    & \textbf{93.47}  & 36.71 &25.701 & 4.434\\ 
\bottomrule
\end{tabular}
\label{tab:Efficient_Compare}
\end{table*}
\par In addition to theoretical analysis, we re-run official implementations and construct controlled variants for empirical validation. FLOPs primarily measure arithmetic complexity, whereas inference latency is jointly influenced by computation, memory access, and hardware utilization. The FLOPs and inference latency of pre-training and classification models are reported in Tab.~\ref{tab:Efficient_Compare}. 
Under input token numbers of 64 and 128, PointER achieves lower FLOPs than Point-MAE and PointMamba, slightly higher FLOPs than Mamba3D. 
In terms of runtime, it consistently exhibits the lowest per-epoch cost during both pre-training and inference.

\par Additionally, the FLOPs of the classification model of PointRWKV are 6.832G with the LGM module and 2.077G with the RWKV module only, indicating that the LGM component in the PRWKV block introduces significantly higher computational overhead compared to the RWKV module. 
Furthermore, PointRWKV (P-RWKV), replacing the PRWKV blocks with the proposed P-RWKV blocks, consistently reduces both computational cost and inference latency, highlighting the efficiency advantage of the proposed design. 
\begin{table}[htbp] 
\caption{\textbf{FLOPs versus input token number.}}
\centering
\scalebox{0.87}{
\begin{tabular}{lccccccc} 
\toprule 
Methods    & 128    & 256   & 512   & 1024 & 2048 & 4096 & 8192\\ \midrule 
(a) Transformer  &5.74 &12.08 & 26.58 & 62.81  & 164.28  & 483.18  & 1584.84  \\ 
(b) Point-MAE  &4.93  &10.13  & 21.45 &47.71 &114.72 &306.73 &- \\ 
(c) RWKV & 4.98 & 9.97   &19.93  &39.86  & 79.73  & 159.45   & 318.90  \\
(d) PointER    & 3.93  &9.99  &19.97  &39.94   &79.88   &159.76  &319.51  \\
\bottomrule
\end{tabular}}
\label{tab:FLOPs_tokens}
\end{table}
\par \textcolor{black}{In Tab.~\ref{tab:FLOPs_tokens}, (a) and (c) illustrate the theoretical computational complexity derived from $FLOPs^{Trans}\approx L((8+4r)Nd^2+4N^2d)$ and $FLOPs^{RWKV}\approx L(A+4r)Nd^2$, respectively. (b) and (d) present the empirical FLOPs of the classification models of Point-MAE and PointER under varying input token numbers. As shown in (d), the empirical results of PointER closely align with the theoretical results in (c), validating the accuracy of the complexity analysis.} 

\subsection{Architectural Generality} 
\subsubsection{PointER-MM} 
To further validate the architectural generality of the proposed modules in multimodal representation learning, we integrate the proposed P-RWKV block into the TAP~\cite{Wang_2023_TAP} framework to construct a multimodal representation learning model, termed PointER-MM. PointER-MM leverages fine-grained supervision from rendered images to enhance point cloud feature representations and is evaluated on downstream point cloud understanding tasks.
\par The embedding module is identical to that of PointER. Subsequently, the encoder extracts features $F_{3D}\in \mathbb{R}^{\hat{g}\times \hat{d}}$ from the input point cloud $P$, where $\hat{g}$ denotes the number of centers and $\hat{d}$ is the feature dimension. 
Then, the camera pose $R^{\alpha}\in \mathbb{R}^{3\times 3}$ of the target rendered images $V^{\alpha}$, together with $F_{3D}$, are fed into the Photograph Module to predict the corresponding image features $F^R_{2D}\in \mathbb{R}^{h\times w\times \hat{d}}$, where $h$ and $w$ denote the spatial resolution of the predicted feature map. 
Finally, a 2D Generator decodes $F^R_{2D}$ into an RGB image $I_{gen}\in \mathbb{R}^{H\times W\times 3}$, where $H$ and $W$ represent the height and width of the generated view. Following TAP, we employ the Mean Squared Error (MSE) loss as the training objective.
\begin{table}[htbp]
\caption{\textbf{Shape classification results of PointER-MM.} 
}
\centering
\scalebox{0.9}{
\begin{tabular}{lcccc}
\toprule
\centering  
Methods      & ModelNet40  & OBJ-BG & OBJ-ONLY & PB-T50-RS \\ \midrule
Point-MAE~(Rep.)    & 93.24 & 90.02   & 88.29   & 85.18 \\
PointER~(Point-MAE)    & \textbf{93.41} & \textbf{90.38}  & 88.49  & 84.93 \\ 
TAP~(Rep.)   & 93.67 & 90.36  & \textbf{89.50}  & 85.67 \\ 
PointER-MM~(TAP)  & \textbf{93.73} & \textbf{90.61}  & 89.44  & \textbf{86.00} \\ 
\bottomrule
\end{tabular}}
\label{tab:shapeCls_PointER_MM}
\end{table}
\par \textbf{Shape classification of PointER-MM.} We evaluate PointER-MM on ModelNet40 and ScanObjectNN, as shown in Tab.~\ref{tab:shapeCls_PointER_MM}. 
Our method achieves performance comparable to TAP~\cite{Wang_2023_TAP}, with a marginal gain of $+0.06\%$. On ScanObjectNN, PointER-MM further outperforms TAP under the more challenging `OBJ-BG' and `PB-T50-RS' settings.
\par These results demonstrate that P-RWKV generalizes well across both single-modality and cross-modality frameworks. Moreover, the learned representations exhibit strong discriminative power for classification.

\subsubsection{PointGLA}
\par To further validate the architectural generality of the proposed modules beyond the P-RWKV architecture, we incorporate LPE and SCE into the Gated Linear Attention (GLA) architecture to derive the P-GLA block. Based on stacked P-GLA blocks, we further build the representation learning framework PointGLA and evaluate it on object-level shape classification tasks.
\begin{figure}[htbp]
    \centering
    \includegraphics[width=0.79\linewidth]{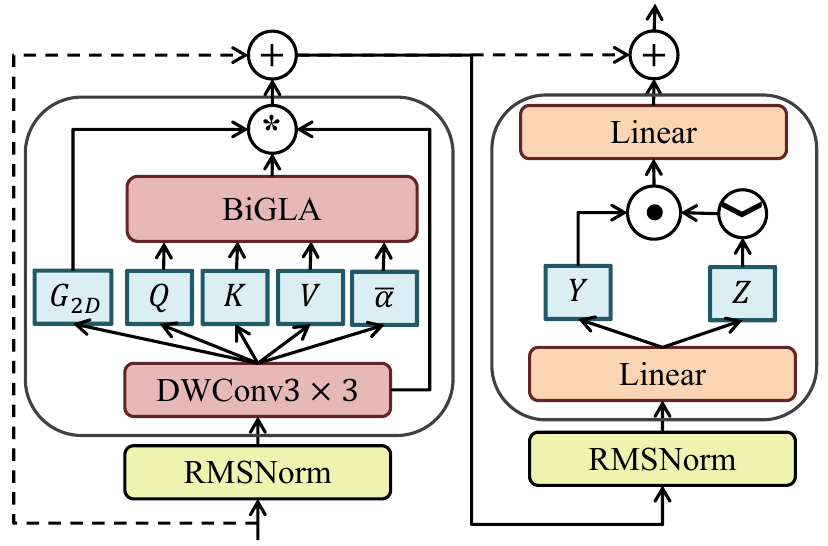} 
    \includegraphics[width=0.85\linewidth]{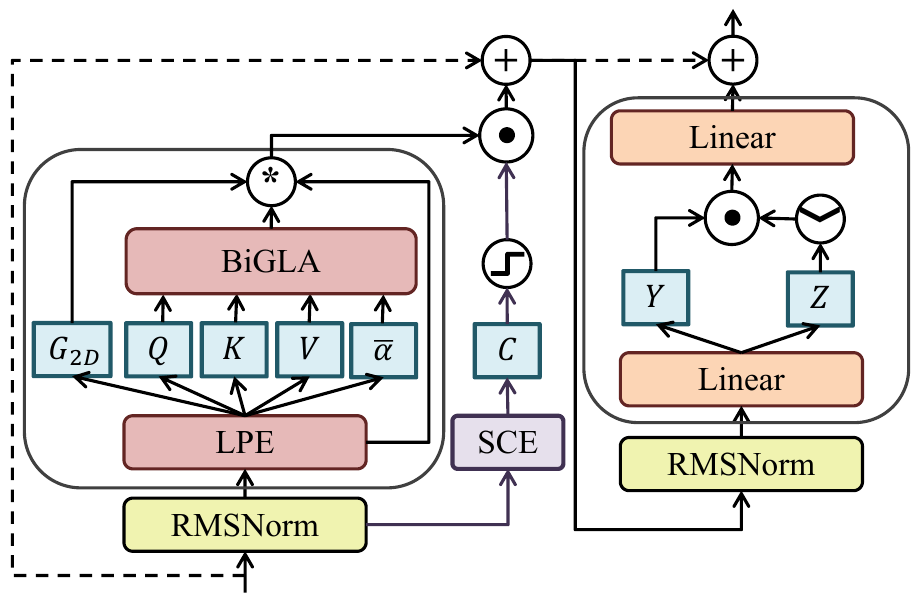} 
    \includegraphics[width=1.0\linewidth]{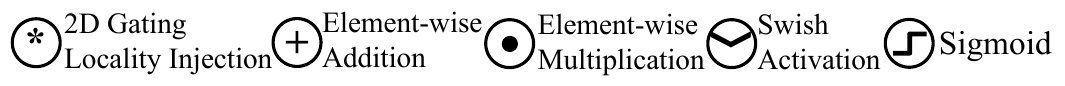}
    \caption{\textbf{Visualization of the ViG block from \cite{ViG_AAAI25} (top) and the proposed P-GLA block (bottom) integrating the LPE and SCE modules.}
    The bidirectional GLA (BiGLA) layer proposed in ViG~\cite{ViG_AAAI25} introduces a directional gating mechanism to adaptively select global contextual information from different directions. 
    Most parameters are shared between the forward and backward directions, enabling efficient bidirectional dependency modeling.
}
    \label{fig:GLA_LPE_SCE}
\end{figure}
\par \textcolor{black}{GLA~\cite{GLA_ICML24} is a hardware-efficient linear attention mechanism that trades off memory movement against parallelizability. The vanilla GLA model was originally developed for NLP, with the details provided in \cite{GLA_ICML24}. 
Subsequently, ViG~\cite{ViG_AAAI25} extends GLA to the image domain and benchmarks it against representative visual backbones, including DeiT~\cite{DeiT_21_ICML}, Swin~\cite{Swin_Transformer_ICCV21}, VMamba~\cite{VMamba_NeurIPS24}, and VRWKV~\cite{Vision_RWKV_ICLR25}. }
\par \textcolor{black}{Inspired by the ViG block~\cite{ViG_AAAI25}~(top row of Fig.~\ref{fig:GLA_LPE_SCE}), we adapt the GLA block to the point cloud domain by seamlessly integrating the proposed LPE and SCE modules, yielding the P-GLA block (bottom row of Fig.~\ref{fig:GLA_LPE_SCE}). 
In this design, LPE replaces the original $DWConv3\times 3$ operator, while SCE follows the spatial-mix formulation in the P-RWKV block and is fused with the BiGLA output via element-wise multiplication. 
Following the destruction–reconstruction pre-training paradigm of Point-MAE, we further construct PointGLA, whose encoder and decoder are composed of the proposed P-GLA blocks. }
\begin{table*}[htbp]
\caption{
\textbf{Inference time and shape classification results of PointGLA.} 
}
\centering
\begin{tabular}{lccccccc}
\toprule
\centering  
Methods                  &Backbone  & Inference E-T~(s)~$\downarrow$ &Param.(M)~$\downarrow$    & ModelNet40~$\uparrow$  & OBJ-BG~$\uparrow$ & OBJ-ONLY~$\uparrow$ & PB-T50-RS~$\uparrow$\\ \midrule 
\textcolor{black}{PointER~(Sup.)}  & RWKV   & 27.15 & 25.7  & 93.34 & 85.54 & 86.41 &82.55\\ 
\textcolor{black}{PointGLA~(Sup.)} & GLA    & 42.04  &22.24  & 93.45 & 84.85 & 85.37 & 84.25 \\ 
\midrule
Point-MAE~(Rep.)     & Transformer   &38.49 &22.1     & 93.24  & 90.02   & 88.29   & 85.18\\
PointER          & RWKV         &27.15  &25.7    & 93.41   & \textbf{90.38}  & 88.49  & 84.93 \\ 
\textcolor{black}{PointGLA}   & GLA     & 42.04  &22.24   & \textbf{93.73} & 90.02 &\textbf{89.98} &\textbf{85.57} \\ 
\bottomrule
\end{tabular}
\label{tab:Abl_MN40_ScanObj_SubMod}
\end{table*}
\par \textbf{Shape classification of PointGLA.} 
As reported in Tab.~\ref{tab:Abl_MN40_ScanObj_SubMod}, PointGLA achieves competitive performance on both datasets under both full-supervised and pre-trained settings. Notably, the fully supervised model attains the best result on `PB-T50-RS', while the pre-trained model further outperforms Point-MAE on both `OBJ-ONLY' and `PB-T50-RS'. 

\begin{table*}[htbp]
\caption{\textcolor{black}{\textbf{Semantic segmentation results on S3DIS.} 
We report the results under the Area 5 and 6-fold cross-validation protocols. $mIoU$ (\%) and $mAcc$ (\%) denote the mean intersection-over-union and mean accuracy, respectively.}}
\centering
\begin{tabular}{lcccccccccccccccc}
\toprule
\multirow{2}{*}{\textbf{Methods}}  & \multicolumn{2}{c}{\textbf{Area 1}} & \multicolumn{2}{c}{\textbf{Area 2}}  & \multicolumn{2}{c}{\textbf{Area 3}} & \multicolumn{2}{c}{\textbf{Area 4}} & \multicolumn{2}{c}{\textbf{Area 5}}  & \multicolumn{2}{c}{\textbf{Area 6}} & \multicolumn{2}{c}{\textbf{6-Fold}}\\ 
\cmidrule(lr){2-3} \cmidrule(lr){4-5} \cmidrule(lr){6-7}  \cmidrule(lr){8-9} \cmidrule(lr){10-11}  \cmidrule(lr){12-13} \cmidrule(lr){14-15} \cmidrule(lr){16-17} 
 & mAcc & mIoU     & mAcc & mIoU     & mAcc & mIoU      & mAcc & mIoU     & mAcc & mIoU     & mAcc & mIoU   & mAcc & mIoU\\ 
\midrule
PTv3~\cite{PointTransV3_CVPR_24}   & 89.92  & 83.01     & 74.44 & 63.42      & 94.45 & 86.66        & 81.11 & 71.34       & 78.92 & 73.43        & 93.55  & 87.31       & 85.31 & 77.70\\
\midrule
PTv3-RWKV             & 90.41  & 83.96        & 76.03 & 64.91  & 95.89 & 87.72    & 83.41 & 73.82       & 80.25  & 74.32     & 94.18  & 89.11    & 86.50  & 79.04\\
\bottomrule
\end{tabular}
\label{tab:SemSeg_Area5_6Fold_PTv3}
\end{table*}

\subsubsection{PTv3-RWKV}
\textcolor{black}{
To further validate the architectural generality of the proposed modules in large-scale scene-level point cloud learning, we integrate the core spatial-mix and channel-mix modules into PTv3~\cite{PointTransV3_CVPR_24} with minimal adaptation, resulting in PTv3-RWKV. Several recent works, including PPT~\cite{PPT_2024CVPR} and Sonata~\cite{Sonata_2025_CVPR}, adopt PTv3 as their baseline. We evaluate PTv3-RWKV on large-scale scene understanding benchmarks to assess the effectiveness of the proposed modules in a fundamentally different architecture.
} 
\begin{figure}[htbp]
 \centering
\includegraphics[width=1.0\linewidth]{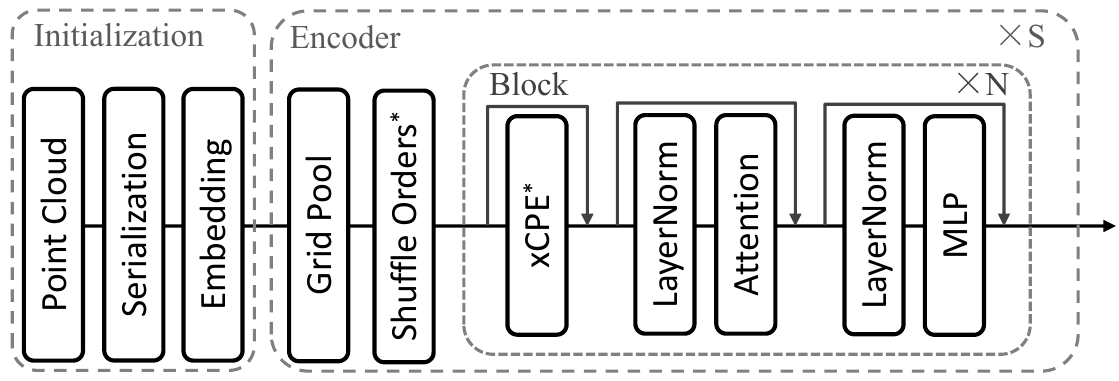}
    \caption{
    \textbf{Overall architecture of PTv3~\cite{PointTransV3_CVPR_24}.}
    } 
    \label{fig:PTv3_OverallArch}
\end{figure}
\par \textcolor{black}{As shown in Fig.~\ref{fig:PTv3_OverallArch}, PTv3~\cite{PointTransV3_CVPR_24} first serializes the point cloud, transforming unstructured points into a structured token sequence. Patch groups are then embedded to form tokens, which are fed into the encoder for hierarchical feature extraction. The encoder is composed of a Grid Pool operation, a Shuffle Order module, and N stacked blocks. Each block integrates an enhanced conditional positional encoding (xCPE) module, two Layer Normalization (LN) layers, a Serialized Attention layer, and an MLP.} 
\begin{figure}[htbp]
 \centering
\includegraphics[width=1.0\linewidth]{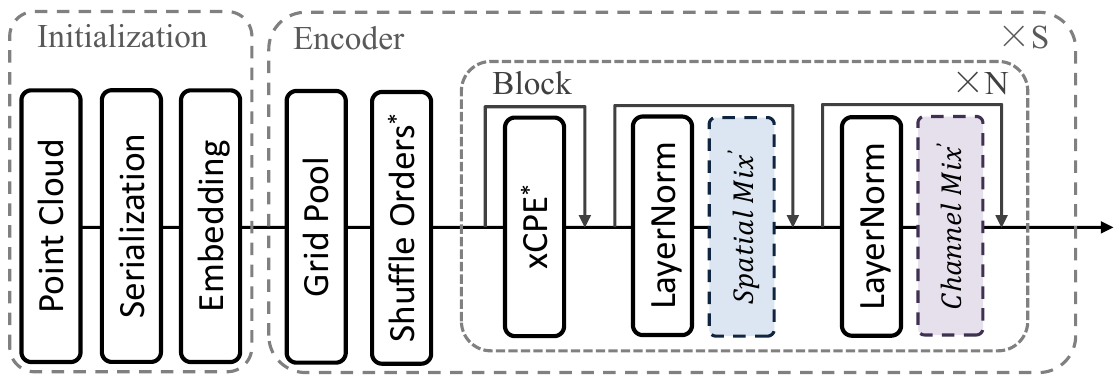}
    \caption{
    \textbf{Overall architecture of the proposed PTv3-RWKV based on PTv3~\cite{PointTransV3_CVPR_24}.}
    } 
    \label{fig:PTv3_RWKV}
\end{figure}
\par \textcolor{black}{We replace the attention and MLP modules in PTv3 with the proposed spatial-mix and channel-mix modules, respectively, with minimal modification, resulting in PTv3-RWKV, as illustrated in Fig.~\ref{fig:PTv3_RWKV}. The implementation will be released together with the PointER codebase.} 

\par \textcolor{black}{Following the experimental settings of PTv3, we conduct semantic segmentation on two real-world datasets (S3DIS~\cite{S3DIS_2016CVPR} and ScanNet v2~\cite{scannet_2017CVPR}).} 
\par \textcolor{black}{\textbf{Semantic segmentation on S3DIS.}} 
Following PTv3, we evaluate semantic segmentation on S3DIS~\cite{S3DIS_2016CVPR} using the Area 5 and 6-fold cross-validation protocol~\cite{qi2017pointnet}. As reported in Tab.~\ref{tab:SemSeg_Area5_6Fold_PTv3}, PTv3-RWKV consistently outperforms the baseline PTv3, indicating that the spatial-mix and channel-mix modules effectively enhance feature extraction and spatial awareness.
\begin{table}[htbp] 
\caption{\textcolor{black}{\textbf{Semantic segmentation results on the ScanNet V2 validation sets.} 
}}
\centering
\begin{tabular}{lccc}
\toprule
\multirow{2}{*}{\textbf{Methods}}   & \multicolumn{3}{c}{\textbf{ScanNet Val}}\\ 
\cmidrule(lr){2-4} 
 & mIoU~(\%) & mAcc~(\%) & oAcc~(\%)  \\ 
\midrule
PTv3~\cite{PointTransV3_CVPR_24}   & 77.5 & 85.0 & 92.0  \\
\midrule
PTv3-RWKV                     & 78.89~(+0.39) & 85.82~(+0.82) & 92.23~(+0.23) \\
\bottomrule
\end{tabular}
\label{tab:SemSeg_PTv3_ScanNetV2}
\end{table}
\par \textcolor{black}{\textbf{Semantic segmentation on the ScanNet v2 validation set.} We further evaluate PTv3-RWKV on another indoor dataset, ScanNet v2, which contains 1,201 training samples, 312 validation samples, and 100 test samples. The model is trained on the training split and evaluated on the validation split, with the results summarized in Tab.~\ref{tab:SemSeg_PTv3_ScanNetV2}. PTv3-RWKV again outperforms the baseline PTv3 across three metrics: $mIoU$, $mAcc$, and $oAcc$.}

\subsection{Ablation study} \label{sec:Ablation}
We conduct ablation studies to analyze the key design choices of PointER and to validate the effectiveness and generality of the proposed modules. 

\subsubsection{Encoder Depth}
We construct pre-training models with a fixed decoder depth of 4 and vary the encoder depth among 8, 12, and 16. The models are pre-trained on ShapeNet55 and evaluated on ModelNet40, enabling us to identify a suitable trade-off between expressiveness and efficiency.
\begin{table}[h]
\caption{\textbf{Ablation studies on encoder depth.} 
}
\centering
\scalebox{0.81}{
\begin{tabular}{lcccccc}
\toprule
Method   & Linear-SVM   & Acc.~(Vote) & PreT. E-T~(s) & Inf. E-T~(s) & FLOPs(G)\\ \midrule
PointER (8)    & 86.85 & 92.71  &104.95 &17.61  &1.685\\ 
PointER (12)    & 88.36 & 93.41 & 125.49  & 27.24 &2.217\\
PointER (16)  & 88.89  & 93.44  &149.69 & 31.75  &2.749\\ 
\bottomrule
\end{tabular}}
\label{tab:abl_Encoder_Depth}
\end{table}
\par As shown in Tab.~\ref{tab:abl_Encoder_Depth}, the model with an encoder depth of 8 yields noticeably lower accuracy than the model with a depth of 12. While increasing the encoder depth to 16 slightly improves the Linear-SVM and voting accuracies over the depth-12 model, this marginal gain comes at a considerable increase in computational cost. Specifically, the pre-training and inference times per epoch increase to approximately 150s and 32s, respectively, compared with 125s and 27s for the depth-12 model. Consequently, we adopt an encoder depth of 12 for our final configuration, as it offers the best balance between accuracy and efficiency.

\subsubsection{Module Design}
We progressively incorporate LPE and SCE into models and evaluate the resulting variants on shape classification (Tab.~\ref{tab:abl_IntroModules_MN40}). 
$Variant_1$ directly applies RWKV to point cloud representation learning without token shifting, while $Variant_2$ introduces temporal token shifting within each block. $Variant_3$ incorporates LPE, and $Variant_4$ further expands its local neighborhood. PointER extends $Variant_3$ with SCE to form the complete P-RWKV block. The results validate the effectiveness of LPE and SCE and suggest that $LPE_2$ is more suitable than $LPE_4$ for object-level classification. 
\par Furthermore, when the input token number is 64, $Variant_2$ requires approximately 111s and 18s per epoch for pre-training and classification, respectively. The corresponding times increase to 119s and 22s for $Variant_3$, and to 126s and 27s for PointER, indicating that the additional overhead introduced by LPE and SCE remains limited.
\begin{table}[htbp]
\caption{
\textbf{Ablation studies on LPE and SCE.} 
}
\centering
\begin{tabular}{lccc}
\toprule
Method         & Backbone            & Acc.   & Acc.~(Vote) \\ \midrule
Point-MAE~(Rep.)   & Transformer    & 92.59  & 93.24    \\ \midrule 
$Variant_1$~(None) & RWKV    & 92.06  & 92.10    \\ 
$Variant_2$~(Original)& RWKV  & 92.54   & 92.83 \\ 
$Variant_3$~($LPE_2$)& RWKV   & 92.71  & 93.29~(+0.46\%) \\  
$Variant_4$~($LPE_4$)& RWKV   & 92.85  & 93.17~(+0.34\%) \\  
$PointER$~($LPE_2$+SCE)& RWKV & 92.93  & 93.41~(+0.58\%)  \\ 
\hline
\textcolor{black}{$Variant_5$~(Original)} & GLA & 92.57  & 93.19 \\ 
\textcolor{black}{$Variant_6$~($LPE_2$)} & GLA & 92.97  & 93.62~(+0.46\%) \\
\textcolor{black}{$PointGLA$~($LPE_2$+SCE)} & GLA & 93.07  & 93.73~(+0.54\%) \\ 
\bottomrule
\end{tabular}
\label{tab:abl_IntroModules_MN40}
\end{table}

\par 
Based on these observations, PointGLA is constructed using only the core $LPE_2$ and SCE modules. The results further indicate that $LPE_2$ contributes more substantially to performance improvement than SCE.

\section{Conclusion} \label{sec:Conculsion}
The quadratic complexity of standard Transformers limits their scalability and efficiency for large-scale point cloud applications. To address this, we propose P-RWKV, an RWKV-based architecture for efficient point cloud representation learning. Through the integration of LPE and SCE, P-RWKV enhances local geometric perception and spatial awareness while preserving linear computational complexity.
\par \textcolor{black}{Beyond validating P-RWKV as a standalone model, we further demonstrate the generality and plug-and-play flexibility of its components through rigorous module-level evaluations across multiple architectures and learning paradigms. 
Extensive experiments show that P-RWKV and its sub-modules achieves performance on par with or superior to transformer- and mamba-based counterparts, particularly on fine-grained tasks, while achieving the lowest inference latency.}

\par \textcolor{black}{Overall, P-RWKV represents a powerful and modular extension of the RWKV family, offering an efficient and flexible alternative for large-scale resource-constrained 3D perception and paving the way for broader adoption of linear-complexity architectures. Future research will explore RWKV-based architectures for large-scale outdoor scene understanding, where efficiency and scalability become increasingly critical.}





 
\bibliographystyle{IEEEtran}
\bibliography{IEEEfull}
 
\vspace{11pt}

\vspace{11pt}

\vspace{11pt}

\vfill

\end{document}